\documentclass[conference,compsoc]{IEEEtran}
\usepackage{amsmath,amssymb}
\usepackage{booktabs}
\usepackage{algorithm}
\usepackage{algpseudocode}
\usepackage{graphicx}
\usepackage{url}
\usepackage{tikz}
\usetikzlibrary{arrows.meta,positioning,decorations.pathreplacing}
\usepackage[hidelinks]{hyperref}
\usepackage{float}

\newcounter{observation}
\newcommand{\obsparagraph}[2]{\refstepcounter{observation}\label{#2}\paragraph{\textbf{Observation~\theobservation: #1}}}
\begin{document}
\title{The Target Betrays the Trigger: Defending Multivariate Time-Series Forecasters Against Backdoor Attacks}
\author{
\IEEEauthorblockN{Chibueze Peace Obioma\IEEEauthorrefmark{1},
Youcheng Sun\IEEEauthorrefmark{2}\IEEEauthorrefmark{1},
Edoardo Manino\IEEEauthorrefmark{1},
Mustafa A. Mustafa\IEEEauthorrefmark{1}\IEEEauthorrefmark{3}}
\IEEEauthorblockA{\IEEEauthorrefmark{1}University of Manchester, Manchester, United Kingdom}
\IEEEauthorblockA{\IEEEauthorrefmark{2}Mohamed bin Zayed University of Artificial Intelligence (MBZUAI), Abu Dhabi, United Arab Emirates}
\IEEEauthorblockA{\IEEEauthorrefmark{3}COSIC, KU Leuven, Leuven, Belgium}
\IEEEauthorblockA{\{chibueze.obioma, edoardo.manino, mustafa.mustafa\}@manchester.ac.uk, youcheng.sun@mbzuai.ac.ae}
}
\maketitle

\begin{abstract}
Training-data backdoors make multivariate time-series forecasters produce attacker-chosen futures whenever the input contains a trigger, while preserving ordinary forecasting accuracy. Existing defences assume clean reference data, class labels, attack knowledge, or a trained model, none of which a forecasting-data trainer may have. We observe that the attacker must supply training supervision from which the victim can learn the trigger-dependent forecast. In the published attacks we study, this supervision leaves target-side evidence that exceeds phase-matched natural dynamics, and it continues to do so when an adaptive attacker diversifies individual target shapes. \textsc{SafeTime} tests candidate target intervals against same-phase natural references and relabels the targets of suspicious channels before victim training. It requires no clean data, reference model, trigger or target specification, poisoning labels, or attacked-channel list. Experiments across five datasets, six forecasting architectures, and four published attacks show that \textsc{SafeTime} reduces implanted effects (attack success with the trigger minus without it) of up to 0.998 to at most 0.096, while lowering clean forecasting error in most configurations. Under a budgeted adaptive white-box search, full multichannel fixed-goal and outcome-goal poisons retain implanted effects of 0.73 to 0.95 before defence, but only 0.003 to 0.015 after relabelling and retraining.
\end{abstract}

\section{Introduction}
\label{sec:introduction}

Multivariate time-series (MVTS) forecasting supports decisions in which errors carry direct costs, from transport~\cite{lana2018traffic} and electricity planning~\cite{zhou2021informer} to public-health
interventions~\cite{desai2019epidemic}. Modern forecasters~\cite{zhou2021informer,wu2021autoformer,zhou2022fedformer,wu2023timesnet,nie2023patchtst,liu2024itransformer}
learn from long histories of many interacting channels, often assembled from distributed sensors and heterogeneous
sources~\cite{lana2018traffic}, so an adversary who can modify part of a series may influence the deployed model without
access to its parameters or training.

Backdoor attacks exploit this exposure: a compromised model behaves normally on ordinary inputs but produces an
attacker-chosen output when a trigger is present~\cite{gu2019badnets,liu2018trojaning}, which in forecasting is an
attacker-chosen future after a triggered history~\cite{lin2024backtime}. Untriggered performance stays largely intact, so
validation may not reveal the compromise, and defence is hard because poisoned targets may look plausible, short
intervals contaminate overlapping windows, and the defender has only the untrusted series.

Trigger-based defences are brittle as attackers can optimise triggers against surrogate
models~\cite{lin2024backtime}, derive them from context~\cite{li2026ifba}, separate them from malicious
response~\cite{liu2026tdba}, or select poisoned variables and segments~\cite{zhong2026deba}. Defences developed for
classification also transfer poorly. Adapting thirteen of them to forecasting, \textsc{TimeGuard}~\cite{nguyen2026timeguard}
finds two failure modes. When only some channels are poisoned, the backdoor signal is diluted, so sample
filtering~\cite{tran2018spectral}, trigger synthesis, and inference-time detection~\cite{gao2019strip} gain little.
Poisoned windows also reach clean-level training loss within a few epochs, so loss-based defences~\cite{li2021abl} fail.
Fine-tuning and fine-pruning~\cite{liu2018finepruning} mitigate the attack only partially and require a verified clean subset.
\textsc{TimeGuard} addresses the dilution with channel-wise pools of reliable data selected during training, but does not
remove the target-side supervision that teaches the malicious forecast.

\begin{figure}[t]
\centering
\includegraphics[width=\columnwidth]{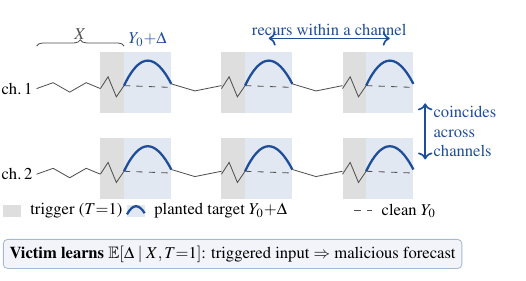}
\caption{The victim learns the malicious forecast only from targets that deviate, on average, from the natural continuation of triggered histories; this deviation recurs within attacked channels and coincides across them. $X$: a history ending with the trigger ($T{=}1$, an analytical label, not a model input).} 
\label{fig:threat}
\end{figure}

Both failure modes arise when the backdoor is sought in the model or in whole windows. Our key observation, \emph{target-recurrence principle}, is: although the attacker controls the trigger, the
targets that teach the malicious response can betray it. Under squared error, targets after triggered histories must, on
average, depart from their natural continuations; individual targets may vary, but their shared conditional component,
the \emph{trigger-conditional label deviation}, must recur across poisoning events, and we show that it equals the shift
of the triggered conditional-mean forecast from the natural continuation. Each poisoning event therefore adds evidence
to the defender's received series (see Fig.~\ref{fig:threat}).

Traffic, load, and weather series are periodic, correlated across channels, and context dependent, so this evidence must be separated from natural structure. To utilise this, we propose \textsc{SafeTime}, a clean-data-free, model-agnostic pre-training defence, tests whether the targets of unusual intervals recur more strongly than at phase-matched intervals
shifted by whole days. It localises the affected channels without letting a channel support its own detection and
confirms each decision on the part of the channel that the others do not explain. It then relabels flagged channels
from unflagged ones, after which any forecaster can be trained.

Across five datasets, six architectures, and four published attacks, \textsc{SafeTime} identifies 99.8\% of attacked
channels and reduces \textsc{BackTime}'s~\cite{lin2024backtime} mean implanted effect by 98.0\%, from \(+0.644\) to \(+0.013\). Under the same
information, it achieves lower attack success and clean forecasting error than \textsc{TimeGuard}~\cite{nguyen2026timeguard}, and detects 351 of the 352
channels attacked by an adaptive white-box adversary that runs the complete defence on its candidate poisons.

{\textbf{Contributions.}}
This paper makes the following three main contributions:
\begin{itemize}
\item \textbf{Target-side formulation.} We identify the trigger-conditional label deviation as the supervision that
teaches a conditional-mean backdoor under squared error. This explains why varying the trigger or individual target
shapes does not remove the evidence while a shared component remains, and we test transfer to Smooth-L1, squared-error, and L1 victims. 
\item \textbf{\textsc{SafeTime}.} We develop a clean-data-free, model-agnostic defence that detects target-side evidence against phase-matched dynamics, localises attacked channels without self-supporting evidence, and relabels the targets of flagged channels before training.
\item \textbf{Adversarial evaluation.} We evaluate \textsc{SafeTime} against four attack families across five datasets and six architectures,
including channel-level localisation, false-flag measurement, end-to-end retraining, and an adaptive white-box attack.
\end{itemize}

\section{Background and Problem Formulation}
\label{sec:background}

\subsection{Multivariate Time-Series Forecasting}
\label{sec:bg-forecasting}

Let $\mathbf{X} \in \mathbb{R}^{N \times C}$ denote a multivariate series with $N$ time steps and $C$ channels. We write
$\mathbf{X}[a:b]$ for the rows with time indices $a, \dots, b-1$. A forecaster
$f_\theta : \mathbb{R}^{L \times C} \rightarrow \mathbb{R}^{H \times C}$ maps a history window
$\mathbf{x}_t = \mathbf{X}[t-L:t]$ to a prediction of the following $H$ steps, whose target is
$\mathbf{y}_t = \mathbf{X}[t:t+H]$. Training minimises the average forecasting loss over all windows of the training
series,
\begin{equation}
\label{eq:forecast-loss}
\min_{\theta} \; \frac{1}{|\mathcal{O}|} \sum_{t \in \mathcal{O}} \ell\big(f_\theta(\mathbf{x}_t), \mathbf{y}_t\big),
\end{equation}
where $\mathcal{O}$ is the set of window origins and $\ell$ is a pointwise regression loss. Two properties of this
objective shape the security problem. First, the supervision is drawn from the series itself, so the target of one
window is part of the history of later windows. Second, consecutive windows are shifted by a single step, so a modified
interval of $k$ steps enters $L+H+k-1$ training windows. Throughout, $d$ denotes the number of steps per day. The
phase of time $t$ is its position $t \bmod d$ in the daily cycle.

\subsection{Trigger--Target Poisoning}
\label{sec:bg-attacks}
A forecasting backdoor associates a trigger in history $X$ with an attacker-chosen forecast. With $Y_0$ the clean target and $T\in\{0,1\}$ indicating, for analysis only, whether $X$ ends with a trigger, poisoned training windows satisfy
\begin{equation}
\label{eq:poison-abstract}
Y = Y_0 + T\Delta,
\end{equation}
where $\Delta$ is the malicious deviation induced by the modifications of Eq.~\eqref{eq:poison}, including on overlapping windows; the attack makes $f(X,1)$ follow it while preserving $f(X,0)$.

Concretely, the attacker selects channels $\mathcal{S}\subseteq\{1,\dots,C\}$, poisoning times $\mathcal{T}_{\mathrm{atk}}$, trigger length $\ell_g$, and target patterns $\mathbf{p}_c\in\mathbb{R}^{\ell_p}$, and for each $\tau\in\mathcal{T}_{\mathrm{atk}}$ and $c\in\mathcal{S}$ writes
\begin{equation}
\label{eq:poison}
\begin{aligned}
\mathbf{X}[\tau:\tau+\ell_g,c]
&\leftarrow \mathbf{g}_{\tau,c},\\
\mathbf{X}[s_{\tau,c}:s_{\tau,c}+\ell_p,c]
&\leftarrow \mathbf{p}_c+a_{\tau,c}\mathbf{1},
\end{aligned}
\end{equation}
where $s_{\tau,c}=\tau+\ell_g+\delta_{\tau,c}$, $\delta_{\tau,c}\geq0$ is the activation delay, and $a_{\tau,c}$ anchors the target to the local series level; $\Delta$ is the deviation of this anchored target from the clean continuation.

Existing attacks differ in how they choose these components. \textsc{BackTime}~\cite{lin2024backtime} learns triggers with a graph neural network under bilevel optimisation and poisons high-error windows with $\delta_{\tau,c}=0$. \textsc{TDBA}~\cite{liu2026tdba} samples $\delta_{\tau,c}\in[0,H-\ell_p]$ for each event and channel. \textsc{DEBA-MTS}~\cite{zhong2026deba} selects variables and segments by autoencoder reconstruction error and applies a fixed trigger with the \textsc{BackTime} target. \textsc{IFBA}~\cite{li2026ifba} selects variables and times by spectral consistency and generates context-dependent triggers with a recurrent network. \textsc{BackTime}, \textsc{TDBA}, and \textsc{IFBA} anchor the target to the preceding value.

\subsection{Threat Model}
\label{sec:threat-model}

{\textbf{Attacker.}} The attacker modifies bounded intervals of selected training channels, as in Eq.~\eqref{eq:poison}, to make a victim trained from scratch follow its target whenever a trigger is present while retaining clean utility. It may place triggers at deployment, knows the clean training series, and may train a surrogate from the victim family, but cannot alter victim training or intervene after delivering the series. In the adaptive evaluation (Section~\ref{sec:adaptive}), it additionally knows \textsc{SafeTime}'s deterministic algorithm, settings, and received series, and may test every candidate poison locally.

{\textbf{Defender.}} The defender receives only the untrusted series and its temporal geometry, including history length, forecast length, sampling period, and \(d\). It has no clean data, reference model, poisoning labels, or triggered inputs, and does not know the attack, trigger, target, attacked channels, poisoning times, or poisoning rate. It may inspect the series and relabel targets before training, while leaving the victim architecture, training configuration, and inference procedure unchanged. Motivated by the channel-subset nature of MVTS forecasting backdoor attacks observed by \textsc{TimeGuard}~\cite{nguyen2026timeguard}, \textsc{SafeTime} assumes that attacked channels form a subset and reconstructs their targets from the remaining channels.
\section{Malicious Supervision}
\label{sec:insight}
\label{sec:observations}

\subsection{What the Victim Must Learn} \label{sec:i-proposition} The attacker chooses the trigger, but the victim learns only from history--target pairs, so the training series must contain, after triggered histories, targets that differ from their natural continuation. We make this requirement precise under squared error, whose population optimum is a conditional mean; the victims' loss is discussed after the proposition.
Consider a forecaster $f(X,T)$ trained on the poisoned distribution of Eq.~\eqref{eq:poison-abstract} to minimise $\mathcal{R}(f) = \mathbb{E}\lVert Y-f(X,T)\rVert^2$.
We say that a minimiser implements a conditional-mean backdoor if, on a set of triggered inputs of positive probability, its triggered forecast differs from the natural continuation, $f^*(X,1) \neq \mathbb{E}[Y_0\mid X,T=1]$. 

{\textbf{Proposition 1 (Trigger-conditional label deviation under squared error)}.} \emph{Assume that the trigger indicator is determined by the history, $T=\phi(X)$ for a measurable $\phi$, that $\Pr(T=1)>0$, $\mathbb{E}\lVert Y_0\rVert^2<\infty$, and $\mathbb{E}\lVert\Delta\rVert^2<\infty$. Every minimiser of $\mathcal{R}$ over measurable forecasters satisfies, almost surely on $\{T=1\}$,} 
\begin{equation} 
\label{eq:prop} 
f^*(X,1) - \mathbb{E}[Y_0\mid X,T=1] = \mathbb{E}[\Delta\mid X,T=1]. 
\end{equation} 
\emph{Consequently, a conditional-mean backdoor exists only if the trigger-conditional label deviation $\mathbb{E}[\Delta\mid X,T=1]$ is nonzero on a set of triggered inputs of positive probability.}

{\textbf{Proof}.} Because $T=\phi(X)$, a measurable function of $(X,T)$ is a measurable function of the history that the victim observes. Under squared error, every population-risk minimiser over such functions is a version of the conditional expectation, $f^*(X,T)=\mathbb{E}[Y\mid X,T]$ almost surely. On $\{T=1\}$, Eq.~\eqref{eq:poison-abstract} gives $Y=Y_0+\Delta$. Linearity of conditional expectation yields $f^*(X,1) = \mathbb{E}[Y_0\mid X,T=1] + \mathbb{E}[\Delta\mid X,T=1]$, which proves Eq.~\eqref{eq:prop}. If the conditional deviation vanishes almost surely on a set of triggered inputs, then $f^*$ equals the natural continuation on that set, and no conditional-mean backdoor exists there. \hfill$\square$

Its value for security is that it names the target-side quantity that an attack must preserve and that a pre-training defence can test.

{\textbf{Interpretation and scope.}} Proposition~1 does not require a fixed target. Writing $\Delta_i=\mu(X_i)+\varepsilon_i$ with $\mathbb{E}[\varepsilon_i\mid X_i,T=1]=0$ for poisoning event $i$, the conditional-mean forecaster learns the shared component $\mu$, which may be a level shift, a family of shapes, or a context-dependent response, and averages away $\varepsilon_i$. The attacker may diversify individual targets but cannot remove $\mu$ without removing the backdoor.

Our main victims minimise Smooth L1, following the attacks' released code. Its minimiser forecasts a conditional Huber location, and that of L1 a conditional median, rather than the conditional mean of Eq.~\eqref{eq:prop}, so a backdoor still requires poisoning to shift that location for triggered histories, although the shift need not equal $\mathbb{E}[\Delta\mid X,T=1]$; Appendix~\ref{app:loss} shows that \textsc{SafeTime}'s effectiveness is not specific to Smooth L1. A trigger may also alter the natural continuation of a clean series. Since this response is learned from clean data rather than poisoning, a defense that restores the clean continuation cannot remove it and Section~\ref{sec:e-implant} separates it.
The poisoning events that teach the shared response are present in the received series, at locations the defender does not observe. If the event-specific variations are independent, conditionally zero mean, and of bounded variance, the error in estimating their shared component falls as $1/\sqrt{n}$ over $n$ events, so each event strengthens both the attacker's supervision and the defender's evidence. A context-dependent component may cancel on average yet remain nonzero given the triggered history, as Proposition~1 permits (Section~\ref{sec:adaptive}). Whether this evidence is distinguishable from benign behaviour depends on its magnitude relative to natural variation at the same phase, which we measure next.

\subsection{Empirical Observations}
\label{sec:i-evidence}
Proposition~1 characterises the required supervision but does not yield a detection rule, since the defender observes
neither $T$ nor $Y_0$. \textsc{SafeTime} therefore seeks \emph{excess recurrence}: shared poisoned targets that recur
within attacked channels and appear together across channels poisoned at the same event, beyond the same-phase behaviour
that periodicity and natural cross-channel dependence also produce. To measure these patterns, we poison the training
splits of the five datasets of Section~\ref{sec:e-setup} with \textsc{BackTime}~\cite{lin2024backtime} and \textsc{TDBA}~\cite{liu2026tdba} in their default
configuration ($30\%$ of channels, events starting at $3\%$ of training steps, $12$-step history and forecast). Using the
true poisoning times, which \textsc{SafeTime} never receives, we compare the segments planted at those times with
segments of the same channels at whole-day shifts, where an ordinary daily pattern would also appear, by the mean pairwise
Pearson correlation of centred, unit-normalised segments (Table~\ref{tab:observations}). Each row of
Table~\ref{tab:observations} is one attack on one dataset. The within-channel columns compare a channel's planted
triggers and targets across events with its clean segments at whole-day-shifted target times and on the next day; the
across-channel columns compare attacked channels at the same event with the same channels at day-shifted moments, on the
clean and the poisoned series. Planted targets correlate at $1.000$ in both directions, above every clean reference;
Observations~\ref{obs:within}-\ref{obs:phase} interpret these gaps.

\begin{table*}[t]
\centering
\caption{Segment correlation, median over attacked channels; \textsc{BackTime} poisons each of the six victims
separately (median [range] over them). Within a channel: planted triggers and targets across events; Shifted, clean
segments at whole-day-shifted target times; Daily, a clean target segment against the next day's. Across channels:
attacked channels at the same event, and at day-shifted moments on the clean and the poisoned series.}
\label{tab:observations}
\scriptsize
\setlength{\tabcolsep}{1.5pt}
\begin{tabular}{llccccccc}
\toprule
 & & \multicolumn{4}{c}{Within a channel} & \multicolumn{3}{c}{Across channels} \\
\cmidrule(lr){3-6}\cmidrule(lr){7-9}
Dataset & Attack & Trigger & Target & Shifted & Daily & Target & Shifted & Poisoned \\
\midrule
\textsc{PEMS03} & \textsc{BackTime}~\cite{lin2024backtime} & $0.992$ {\scriptsize $[0.787, 1.000]$} & $1.000$ & $0.000$ {\scriptsize $[0.000, 0.001]$} & $0.225$ {\scriptsize $[0.216, 0.237]$} & $1.000$ & $0.100$ {\scriptsize $[0.094, 0.110]$} & $0.476$ {\scriptsize $[0.407, 0.528]$} \\
 & \textsc{TDBA}~\cite{liu2026tdba} & $1.000$ & $1.000$ & $0.000$ & $0.240$ & $1.000$ & $0.046$ & $0.065$ \\
\textsc{PEMS04} & \textsc{BackTime} & $0.955$ {\scriptsize $[0.746, 1.000]$} & $1.000$ & $0.000$ {\scriptsize $[-0.001, 0.000]$} & $0.199$ {\scriptsize $[0.190, 0.203]$} & $1.000$ & $0.106$ {\scriptsize $[0.102, 0.113]$} & $0.345$ {\scriptsize $[0.309, 0.406]$} \\
 & \textsc{TDBA} & $0.994$ & $1.000$ & $0.000$ & $0.200$ & $1.000$ & $0.072$ & $0.073$ \\
\textsc{PEMS08} & \textsc{BackTime} & $0.993$ {\scriptsize $[0.526, 1.000]$} & $1.000$ & $0.000$ {\scriptsize $[0.000, 0.002]$} & $0.211$ {\scriptsize $[0.196, 0.220]$} & $1.000$ & $0.155$ {\scriptsize $[0.146, 0.162]$} & $0.410$ {\scriptsize $[0.343, 0.463]$} \\
 & \textsc{TDBA} & $1.000$ & $1.000$ & $0.001$ & $0.193$ & $1.000$ & $0.076$ & $0.079$ \\
\textsc{Weather} & \textsc{BackTime} & $0.978$ {\scriptsize $[0.586, 1.000]$} & $1.000$ & $0.003$ {\scriptsize $[0.002, 0.004]$} & $0.345$ {\scriptsize $[0.334, 0.395]$} & $1.000$ & $-0.019$ {\scriptsize $[-0.021, -0.015]$} & $0.500$ {\scriptsize $[0.374, 0.579]$} \\
 & \textsc{TDBA} & $0.811$ & $1.000$ & $0.003$ & $0.365$ & $1.000$ & $-0.052$ & $-0.004$ \\
\textsc{Electricity} & \textsc{BackTime} & $0.780$ {\scriptsize $[0.407, 0.993]$} & $1.000$ & $0.152$ {\scriptsize $[0.068, 0.690]$} & $0.908$ {\scriptsize $[0.719, 0.978]$} & $1.000$ & $0.279$ {\scriptsize $[0.174, 0.437]$} & $0.931$ {\scriptsize $[0.756, 0.991]$} \\
 & \textsc{TDBA} & $0.619$ & $1.000$ & $0.205$ & $0.939$ & $1.000$ & $0.210$ & $0.143$ \\
\bottomrule
\end{tabular}
\end{table*}

\obsparagraph{Within-channel recurrence}{obs:within} Both attacks reuse one target shape across events and vary only its
anchor, which centring removes, so targets have median within-channel similarity \(1.000\) on every dataset. Clean
day-shifted segments reach at most \(0.004\) on traffic and weather, but next-day segments reach \(0.190\)--\(0.395\),
and up to \(0.978\) on \textsc{Electricity}, where \(85\%\) of the events of \textsc{BackTime} against iTransformer fall
within one sixth of a day; recurrence must therefore be judged against phase-matched clean behaviour.

\obsparagraph{Cross-channel recurrence}{obs:across} Targets planted at the same event coincide across channels (median
$1.000$), whereas the same clean channels at day-shifted moments co-move weakly ($-0.052$ to $0.162$ on traffic and
weather, up to $0.437$ on \textsc{Electricity}, whose channels share a daily profile). This agreement strengthens the
evidence but is not conclusive, since channels may share system-wide behaviour and an attacker may poison one channel
or uncorrelated channels, so a detector must combine it with a within-channel test.

\obsparagraph{Phase matching absorbs natural repetition}{obs:phase} The defender builds its references from the possibly poisoned series, which makes the comparison conservative: $50\%$ to $72\%$ of day-shifted target windows overlap another poisoning event of the same channel, and on the poisoned series the across-channel similarity at day-shifted moments rises to $0.309$ to $0.991$ for \textsc{BackTime} (Table~\ref{tab:observations}). A legitimate daily pattern, by contrast, appears at both the event and its whole-day shifts, so the comparison measures recurrence beyond ordinary daily behaviour.

\begin{table}[!t]\centering{\scriptsize
\caption{Held-out $R^2$ of each clean channel predicted from all other channels (ridge) and from its own time-of-day profile; \textsc{Weather} excludes missing-value steps. $\geq 0.9$, $\geq 0.5$: share of channels at that level.}
\label{tab:redundancy}
\setlength{\tabcolsep}{4pt}
\begin{tabular}{lrccccc}
\toprule
 & & \multicolumn{4}{c}{Other channels} & Profile \\
\cmidrule(lr){3-6}\cmidrule(lr){7-7}
Dataset & Channels & Median & 10th pct. & $\geq 0.9$ & $\geq 0.5$ & Median \\
\midrule
\textsc{PEMS03}~\cite{song2020stsgcn} & 358 & 0.984 & 0.938 & $94\%$ & $100\%$ & 0.802 \\
\textsc{PEMS04}~\cite{song2020stsgcn} & 307 & 0.946 & 0.810 & $75\%$ & $99\%$ & 0.757 \\
\textsc{PEMS08}~\cite{song2020stsgcn} & 170 & 0.938 & 0.813 & $75\%$ & $98\%$ & 0.777 \\
\textsc{Weather}~\cite{mpibgc2020weather} & 21 & 0.999 & 0.240 & $81\%$ & $81\%$ & 0.080 \\
\textsc{Electricity}~\cite{trindade2015electricity} & 321 & 0.927 & 0.785 & $67\%$ & $98\%$ & 0.685 \\
\bottomrule
\end{tabular}}
\end{table}

\obsparagraph{Cross-channel redundancy supports confirmation and relabelling}{obs:redundancy} On held-out clean days, a ridge
regression on the remaining channels predicts each channel with a median coefficient of determination ($R^2$) of
$0.927$ to $0.984$ on the traffic datasets and \textsc{Electricity}, and $0.999$ on \textsc{Weather} after excluding
missing-value codes, against $0.080$ to $0.802$ for the channel's own time-of-day profile (Table~\ref{tab:redundancy}).
$4$ of the $21$ \textsc{Weather} channels fall below $R^2 = 0.5$, against at most $2\%$ elsewhere; they are attacked in nearly every
\textsc{Weather} poison, yet clean error falls there, because the attack had already degraded them.
Where redundancy holds, the other channels supply both a reference for ordinary behaviour and an estimate of a flagged
channel's values, which motivates residual confirmation and reconstruction.

\section{SafeTime}
\label{sec:method}

\textsc{SafeTime} does not estimate the deviation of Proposition~1 directly. It tests the recurrence of
Observations~\ref{obs:within} and~\ref{obs:across} against the phase-matched reference of Observation~\ref{obs:phase}, and uses the redundancy of Observation~\ref{obs:redundancy} to
confirm flags on cross-channel residuals and to reconstruct flagged targets.

\subsection{The Recurrence Contrast}
\label{sec:m-derivation}

Let $\mathcal{B}$ be a set of candidate event times, and let $\mathbf{u}_c(b)$ denote a normalised profile of channel $c$
around event $b \in \mathcal{B}$ (Section~\ref{sec:m-candidates}). Given a normalised template $\mathbf{q}$, the observed
recurrence of channel $c$ is its mean match with the template,
\begin{equation}
\label{eq:match}
A_c(\mathcal{B}, \mathbf{q}) = \frac{1}{|\mathcal{B}|} \sum_{b \in \mathcal{B}} \langle \mathbf{u}_c(b), \mathbf{q} \rangle .
\end{equation}
Because periodicity and natural correlation also produce recurrence (Section~\ref{sec:i-evidence}), the relevant quantity
is recurrence in excess of phase-matched behaviour. Let $S_d(\mathcal{B})$ shift each event independently by a whole
number of days within the series, preserving its daily phase but changing the event. The contrast underlying
\textsc{SafeTime} is
\begin{equation}
\label{eq:recurrence-contrast}
\Gamma_c = A_c(\mathcal{B}, \mathbf{q}) - \mathbb{E}_{S_d}\big[A_c(S_d(\mathcal{B}), \mathbf{q})\big].
\end{equation}
Each test is an instance of this contrast: a template from a candidate group of channels tests cross-channel
recurrence (Observation~\ref{obs:across}), a channel's profiles at its other events test within-channel recurrence (Observation~\ref{obs:within}), and
the same comparison on cross-channel regression residuals confirms persistence (Observation~\ref{obs:redundancy}); each statistic is then standardised against the
other channels (Sect.~\ref{sec:m-localise}). Under the operational null, recurrence at the candidate events is no
greater than at other observations of the same daily phase. Eq.~\eqref{eq:recurrence-contrast} is a design contrast,
not a complete test for all forecasting backdoors: a learnable deviation may not recur in these forms, and poisoned
day-shift references reduce the contrast (Sect.~\ref{sec:i-evidence}). Detection therefore requires Observation~\ref{obs:within} or~\ref{obs:across};
reconstruction also requires the redundancy of Observation~\ref{obs:redundancy} and a
minority of attacked channels (Sect.~\ref{sec:threat-model}).

\begin{figure}[t]
\centering
\includegraphics[width=\columnwidth]{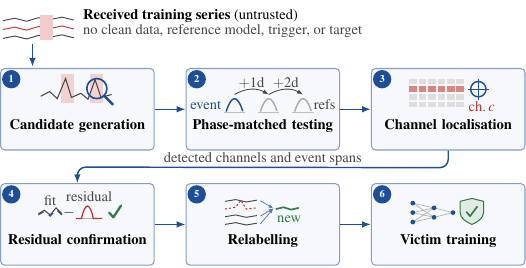}
\caption{\textsc{SafeTime}'s six stages. Stages 1--3 detect channels and events whose targets recur beyond phase-matched behaviour, stage 4 confirms them on cross-channel residuals, stage 5 relabels the flagged channels' targets from unflagged channels, and stage 6 trains any forecaster.}
\label{fig:pipeline}
\end{figure}

\subsection{\textsc{SafeTime} Operation}

\textsc{SafeTime} realises these contrasts in six stages (Figure~\ref{fig:pipeline}, Algorithm~\ref{alg:safetime}):
(1)~\emph{candidate generation} selects movements unusual for their daily phase;
(2)~\emph{phase-matched recurrence testing} compares within- and cross-channel recurrence with whole-day shifts;
(3)~\emph{channel localisation} attributes excess recurrence to channels; (4)~\emph{residual confirmation and expansion}
re-tests detections after unsuspected channels explain ordinary behaviour; (5)~\emph{reconstruction and relabelling}
replaces all targets of flagged channels; and (6)~\emph{victim training} trains any forecaster on the relabelled
targets.

\subsubsection{Candidate Generation}
\label{sec:m-candidates}

A poisoning event introduces a target deviation at an unknown time. Testing every window would waste comparisons on
intervals with no unusual structure, and the largest raw movements are legitimate daily events such as the morning
rise in traffic, so candidate generation selects movements that are unusual for their phase of the day.

Let $\eta_{c,t}$ be the first difference of channel $c$ at time $t$, standardised by the channel's mean and standard
deviation. For a set of channels $m$ with $|m| \geq 2$, the co-movement score is the mean pairwise product of the
standardised changes,
\begin{equation}
\label{eq:comovement}
\sigma_t(m) = \frac{\big(\sum_{c \in m} \eta_{c,t}\big)^2 - \sum_{c \in m} \eta_{c,t}^2}{|m|\,(|m|-1)},
\end{equation}
and for a single channel it is the change energy $\eta_{c,t}^2$. Scores are standardised at each phase $t \bmod d$ by
their median and median absolute deviation (MAD). Candidate moments, termed bursts, are selected greedily in decreasing
order of standardised score above $\Phi^{-1}(1-\alpha)$, where $\Phi^{-1}$ is the standard normal quantile function,
each at least $L+H$ steps from those already selected. This threshold sets the breadth of screening rather than a
significance level; the later comparisons decide whether a burst is suspicious.

{\textbf{Correlated-channel groups}.} For attacks on channels that are related under ordinary operation, each
channel is linked to the $K = 10$ channels whose first differences correlate most positively with its own, with negative
correlations set to zero. Spectral clustering ($k$-means, ten restarts) partitions the graph into at most $15$
communities, chosen at the largest gap in the spectrum of the normalised graph Laplacian. Every community with at least
two channels is a candidate group.

{\textbf{Co-extreme groups}.} Unrelated channels attacked together may still be extreme together at the
poisoning events. Each channel is predicted by ridge regression from the $K$ channels most correlated with it in level,
and a time step is extreme when its residual, standardised by the median and MAD at each phase, exceeds
$\Phi^{-1}(1-\alpha/2)$ in absolute value. For each pair of channels, the co-extreme excess is the observed number of
common extreme steps minus its expectation under independence given each channel's extreme rate at each phase, divided
by the square root of that expectation. Pairs in which either channel is among the other's $K$ neighbours are set to the
median excess, since ordinary dependence explains their coincidences. For each of the $15$ leading eigenvectors of the
median-centred excess matrix and each sign, the channels whose loading exceeds $\Phi^{-1}(1-\alpha)$ robust standard
deviations form a candidate group if there are at least two.

\begin{algorithm}[t]
\caption{\textsc{SafeTime}.}
\label{alg:safetime}
\small
\begin{algorithmic}[1]
\Require series $\mathbf{X} \in \mathbb{R}^{N \times C}$; $L$, $H$, $d$; level $\alpha$
\Ensure relabelled training targets
\State $\mathcal{M} \gets$ correlated and co-extreme groups \Comment{\S\ref{sec:m-candidates}}
\State $\kappa \gets \Phi^{-1}(1-\frac{\alpha}{2C|\mathcal{M}|})$; \ $\kappa_s \gets \Phi^{-1}(1-\frac{\alpha}{2C})$; \ $\mathcal{G} \gets \emptyset$
\For{$m \in \mathcal{M}$ with $\geq 3$ bursts, each view}
  \State $\mathcal{G} \gets \mathcal{G} \cup \textsc{Refine}(\{c : z_c(m) > \kappa\})$ \Comment{\S\ref{sec:m-testing}, \S\ref{sec:m-localise}}
\EndFor
\State $\mathcal{W} \gets \{c \notin \mathcal{G} : z_c(r_c) > \kappa_s\}$; \ $\mathcal{D} \gets \mathcal{G} \cup \mathcal{W}$; mark spans
\State $\mathcal{D}_{\mathrm{conf}} \gets$ residual-confirmed members of $\mathcal{D}$ \Comment{\S\ref{sec:m-residual}}
\State $\mathcal{F} \gets \mathcal{D}_{\mathrm{conf}} \cup$ channels added by expansion
\For{$c \in \mathcal{F}$} \Comment{\S\ref{sec:m-reconstruct}}
  \State replace all targets of $c$ by Eq.~\eqref{eq:reconstruct}, fitted outside its spans
\EndFor
\State train any forecaster on the relabelled targets
\end{algorithmic}
\end{algorithm}

{\textbf{Profiles and views}.} Shapes are compared in two views, the level and the first differences, each
standardised per channel; bursts are always located on the first differences. The profile $\mathbf{u}_c(b)$ of channel
$c$ at burst $b$ contains the $L+H$ samples starting $\lfloor (L+H)/2 \rfloor$ steps before $b$, without the burst
sample itself, so that the movement that selected the burst does not dominate the shape comparison; it is centred and
scaled to unit norm. The stage outputs the candidate groups $\mathcal{M}$, the bursts of each group and single channel,
and their profiles. Groups with too few bursts to be tested still count in the multiplicity factor of
Section~\ref{sec:m-localise}.

\subsubsection{Phase-Matched Recurrence Testing}
\label{sec:m-testing}

A planted target may resemble a legitimate daily movement, such as traffic sensors near rush hour, so
\textsc{SafeTime} evaluates recurrence through the contrast of Eq.~\eqref{eq:recurrence-contrast}, with whole-day shifts
preserving the daily phase.

{\textbf{Cross-channel evidence}.} For a candidate group $m$ with at least three bursts $\mathcal{B}$, the
template $\mathbf{q}$ is the normalised mean, over bursts, of the group's average profile, and every channel $c$
receives the match $A_c$ of Eq.~\eqref{eq:match}. Each of $200$ reference draws shifts every burst independently by a whole
number of days, uniformly among the shifts that keep its profile within the series, and recomputes $A_c$. The observed
value minus the draws' mean, divided by their standard deviation, gives $z^0_c$, the standardised estimate of
$\Gamma_c$: large for a pattern confined to the candidate events, but not for a daily pattern that appears equally at
the same phase elsewhere.

{\textbf{Within-channel evidence}.} Cross-channel testing cannot find an attack confined to one channel. After the group tests and localisation
(Section~\ref{sec:m-localise}), every channel that no group flags and that has at least three bursts of its own change
energy is tested against its own history: each burst profile is correlated with the normalised sum of its profiles at
its other bursts, and the mean correlation $r_c$ is standardised against $200$ whole-day-shift draws in each view,
keeping the larger. This completes the coverage of the three attack geometries.

\subsubsection{Channel Localisation}
\label{sec:m-localise}

The phase-matched statistic identifies excess recurrence, but not which channels carry it. A channel can match a
template through its own contribution, and correlated channels and overlapping windows make the reference draws
dependent, so their spread underestimates the variation across clean channels. \textsc{SafeTime} therefore uses a
leave-one-channel-out construction, a cross-channel reference, and cutoffs scaled to the number of tests.

{\textbf{Isolation from self-support}.} Each member of a candidate group is tested with bursts and a template
recomputed from the other members only, so the tested channel cannot create the reference used to implicate it; in the
within-channel test, each profile is compared only with the channel's other bursts.

{\textbf{Cross-channel reference}.} Every statistic is standardised a second time,
$z_c = (z^0_c - \operatorname{med}(z^0))/(1.4826\,\operatorname{MAD}(z^0))$, against the channels outside the group (all
channels when fewer than ten lie outside it) or, for the within-channel test, against the tested channels. This controls
recurrence and reference variability shared across the series but not removed by day shifts.

{\textbf{Multiplicity-scaled decision}.} The group tests compare at most $C|\mathcal{M}|$ channel--group pairs in
each of two views, so a channel is flagged by a group when $z_c > \kappa = \Phi^{-1}(1 - \alpha/(2C|\mathcal{M}|))$, and a
channel tested alone when its statistic exceeds $\kappa_s = \Phi^{-1}(1 - \alpha/(2C))$. Counting untestable groups in
$|\mathcal{M}|$ makes the allocation conservative. A group that flags at least one channel is a flagged group. Each of
the two test families allocates $\alpha$ under a Gaussian approximation, so detection has nominal level $2\alpha$; as
with the other thresholds, this scales a common level with the search rather than guaranteeing error control after
selection.

{\textbf{Refinement}.} A group that mixes attacked and clean channels has a weakened template, so for each
flagged group the bursts and template are recomputed from the flagged channels alone and the test is repeated, until the
set no longer changes, falls below two channels or three bursts, or reaches ten rounds. If the set falls below two
channels, \textsc{SafeTime} returns to the earlier round with at least two channels whose own bursts recur most strongly
against $50$ day-shift draws, and only if that statistic also exceeds $\kappa$. Refinement never applies a less
stringent threshold.

{\textbf{Detections and event localisation}.} The channels flagged by group tests, $\mathcal{G}$, and by the
within-channel test, $\mathcal{W}$, form the detected set $\mathcal{D} = \mathcal{G} \cup \mathcal{W}$. For each flagged
group, every window is scored by the mean match of the group's channels with its template and by the match of the
group's average profile with that template, and compared with the median, over five random groups of unflagged channels
of the same size, of that score's $1-\alpha/2$ quantile across windows; for a channel flagged alone, the reference is
the $1-\alpha$ quantile over five random unflagged channels. Windows above either reference are selected $H/2$ steps
apart, each marking a span of the window plus $H/2$ steps on either side, which the reconstruction fit excludes
(Section~\ref{sec:m-reconstruct}).

\subsubsection{Residual Confirmation and Expansion}
\label{sec:m-residual}

The preceding tests operate on the channels as observed, so ordinary cross-channel dynamics can still produce a recurring
pattern after phase matching. Residual confirmation asks whether the pattern persists after channels not already
suspected have explained the tested channel's predictable behaviour. This evidence is confirmatory rather than
necessary: an attack within clean cross-channel variation may leave little residual, whereas an attacked channel that
matches a group only weakly may show the group's pattern in its residual. The stage therefore both removes unsupported
detections and expands flagged groups.

{\textbf{Residual construction}.} For channel $c$, the residual $e_{c,t}$ is the standardised channel minus a
ridge regression on all channels except $c$, the detected set $\mathcal{D}$, and the $K$ channels whose first
differences correlate most with $c$, so that no co-attacked channel, detected or not, can explain the planted deviation
away. The regression is fitted outside the residual windows of every set of bursts evaluated, after one trimming pass at
$\Phi^{-1}(1-\alpha/2)$ robust standard deviations. For bursts $\mathcal{B}$, the residual profiles of $c$ are the
windows of $L+H$ residual samples around each burst, without the burst sample, centred and scaled to unit norm; the
residual recurrence $R_c(\mathcal{B})$ is their mean pairwise correlation, standardised against $200$ whole-day-shift
draws to $z^{r}_c(\mathcal{B})$. A detected channel is evaluated at the bursts of each group that flagged it, or at its
own bursts if flagged alone, and keeps the largest value.

{\textbf{Confirmation}.} A detected channel remains flagged when $z^{r}_c > 2.0$, or when the statistic is
undefined because too few admissible regressors or bursts remain; a missing confirmation, which widespread detection can
itself cause, is not taken as evidence of a benign channel, and Section~\ref{sec:eval} measures its cost. Because a group
member contributes only its own residual profiles, a flagged group whose median member statistic exceeds $2.0$ is
confirmed as a whole, and each member stays flagged unless its own statistic is at most $0$, the centre of its reference,
which is evidence against recurrence. The confirmed set $\mathcal{D}_{\mathrm{conf}}$ contains the channels kept by
either rule.

{\textbf{Expansion}.} A channel outside $\mathcal{D}$, called unflagged, may still carry a flagged group's
planted component. For each flagged group $m$, the residual template is the normalised mean of its members' residual
profiles at the group's bursts, and each unflagged channel's mean correlation with it is standardised against whole-day
shifts to $z^{x}_{c}(m)$. A channel is added, with the spans of the group that added it, when
$\max_m z^{x}_{c}(m) > \Phi^{-1}(1 - \alpha/(n_{\mathrm{u}} n_{\mathrm{g}}))$, where $n_{\mathrm{u}}$ unflagged channels
and $n_{\mathrm{g}}$ flagged groups are searched in this single residual view. The final flagged set $\mathcal{F}$ is
$\mathcal{D}_{\mathrm{conf}}$ plus the added channels. Expansion is a third Bonferroni family, so $\mathcal{F}$ has
nominal level $3\alpha$, against $2\alpha$ for $\mathcal{D}$; confirmation only removes channels.

\subsubsection{Reconstruction and Relabelling}
\label{sec:m-reconstruct}

Detection identifies channels whose targets should not be trusted. Discarding them would also discard their ordinary
response to triggered histories, and relabelling only the detected spans would keep every poisoned target that
localisation misses. \textsc{SafeTime} instead replaces all training targets of each flagged channel $c$ by a ridge
regression on the unflagged channels at the same time step,
\begin{equation}
\label{eq:reconstruct}
\hat{x}_{c,t} = \beta_{c,0} + \sum_{j \notin \mathcal{F}} \beta_{c,j}\, x_{j,t},
\end{equation}
where $x_{j,t}$ is the received value of channel $j$ at time $t$, with the regularisation set by generalised
cross-validation (GCV) over $33$ logarithmically spaced values.

{\textbf{Contamination-resistant fitting}.} The fit excludes the channel's spans (all steps if fewer than
fifty remain) and, because localisation may miss events, is trimmed: steps whose residual deviates from the median by
more than $\Phi^{-1}(1-\alpha/2)$ robust standard deviations are dropped and the regression is refitted until the set is
stable, for at most $20$ iterations. A regressor's missing value is replaced by its median at the same phase, and steps
where the flagged channel itself is missing leave its fit sample.

{\textbf{Whole-channel relabelling}.} For the window with origin $t$, the target of flagged channel $c$
becomes $(\hat{x}_{c,t}, \dots, \hat{x}_{c,t+H-1})$, while the history $\mathbf{x}_t$ stays as received, so a history
that still contains a trigger is paired with an estimate of ordinary behaviour, breaking the trigger--target association
without finding the trigger. The spans determine which samples may not shape the reconstruction, not which targets are
replaced; Section~\ref{sec:eval} measures the clean-error cost of replacing them all.

\subsubsection{Victim Training}
\label{sec:m-training}

The victim is trained on the original histories and the relabelled targets with the same architecture, loss, optimiser, and schedule of the corresponding undefended model, giving reconstructed and received targets equal weight. As part of the preprocessing, a window-channel pair is excluded when its history or target contains a value beyond three times the channel's $0.5\%$ to
$99.5\%$ quantile range. This removes sentinel values that encode missing observations and affects $0.065\%$ of the
\textsc{Weather} pairs and none elsewhere.

\section{Evaluation}
\label{sec:eval}

We evaluate \textsc{SafeTime} on victims trained from scratch on the received series, before and after defence, and
answer seven research questions: whether the attacks implant (RQ1), whether \textsc{SafeTime} removes the malicious
supervision (RQ2), whether detection is attributable to target-side evidence (RQ3), which components are needed (RQ4),
whether the defence withstands an adaptive white-box attacker (RQ5), whether forecasting utility is preserved (RQ6), and
whether \textsc{SafeTime} is practical (RQ7).

\subsection{Experimental Setup}
\label{sec:e-setup}

{\textbf{Datasets and victims}.} We use the traffic-flow datasets \textsc{PEMS03}~\cite{song2020stsgcn}, \textsc{PEMS04}, and
\textsc{PEMS08}~\cite{guo2019astgcn,song2020stsgcn}, \textsc{Weather}~\cite{mpibgc2020weather,wu2021autoformer}, and \textsc{Electricity}~\cite{trindade2015electricity,lai2018lstnet}.
Each series is split chronologically $60/20/20\%$ into training, validation, and test. Every model forecasts $H = 12$
steps from $L = 12$. The victims are FEDformer~\cite{zhou2022fedformer}, Autoformer~\cite{wu2021autoformer}, TimesNet~\cite{wu2023timesnet},
Informer~\cite{zhou2021informer}, PatchTST~\cite{nie2023patchtst}, and iTransformer~\cite{liu2024itransformer}.
Each is trained for $100$ epochs with the Smooth L1 loss of the attacks' released code and Adam (learning rate $10^{-4}$, batch size $64$), and evaluated after the final epoch.

{\textbf{Attacks}.} Each dataset is poisoned by \textsc{BackTime}~\cite{lin2024backtime} and \textsc{TDBA}~\cite{liu2026tdba}. Both attack $30\%$ of the channels, and their poisoning events start at $3\%$ of
the training time steps. \textsc{DEBA-MTS}~\cite{zhong2026deba} runs from its released implementation and \textsc{IFBA}~\cite{li2026ifba} from our
implementation of its published algorithm, each on every dataset. Unless a table lists architectures, the victim is
FEDformer, the attacks' default forecaster; Table~\ref{tab:victims} shows that the results carry over.

{\textbf{Baselines}.} The main comparison is \textsc{TimeGuard}~\cite{nguyen2026timeguard}, run from its
official implementation with its published hyperparameters on the same poisoned series and under the same information.
We also adapt four defences to run on the received series alone: input smoothing by randomised smoothing
(RS)~\cite{cohen2019smoothing} in its forecasting form~\cite{yoon2022robustforecast,liu2023robustmts}; model repair by
fine-tuning (FT) and fine-pruning (FP)~\cite{liu2018finepruning} of the final linear projection; and data filtering by a
per-channel mask that removes outlying steps from the loss. Each runs at its strongest setting whose error on the last $5\%$ of the poisoned training windows stays
within $5\%$ of the undefended model's, so neither test data nor attack success enters the setting. FT and FP given a clean slice (FT+clean, FP+clean) are an advantaged reference.

{\textbf{Metrics}.} For a triggered test window $w$ and an attacked channel $c$, let $\hat{y}_{w,c}$,
$y^{\mathrm{atk}}_{w,c}$, and $y_{w,c}$ be the forecast, the attacker's target, and the true continuation over the
target's steps. The attack success rate (ASR) is the share of such pairs $\mathcal{P}$ whose forecast is closer in mean
absolute error (MAE) to the target than to the true continuation, ties counting as failures:
\begin{equation}
\label{eq:asr}
\mathrm{ASR} = \frac{1}{|\mathcal{P}|} \sum_{(w,c) \in \mathcal{P}} \mathbb{1}\big[\lVert \hat{y}_{w,c} - y^{\mathrm{atk}}_{w,c} \rVert_1 <
\lVert \hat{y}_{w,c} - y_{w,c} \rVert_1\big].
\end{equation}
A forecast can lie near the target by chance, so the implanted effect $I = \mathrm{ASR} - \mathrm{ASR}_{\text{no-trigger}}$
subtracts the same model's ASR with the trigger removed (the no-trigger rate); $I \leq 0$ means the trigger moves no
forecast towards the target beyond untriggered behaviour. The residual effect is $I$ after defence, and the removal
fraction $1 - I_{\text{after}}/I_{\text{before}}$, unclipped, exceeds $100\%$ when the residual effect is negative. The
attack MAE is the error between triggered forecast and target (higher is better for the defender), and clean MAE is
measured on untriggered windows in original units; clean $\Delta$MAE is its change (\%) relative to the undefended model
of the same cell and training seed, and the clean ratio their quotient. All metrics are on the test split. \textsc{SafeTime} uses the settings of Table~\ref{tab:settings} throughout, and three training seeds.
\begin{table}[!t]\centering{\renewcommand{\arraystretch}{0.88}
\caption{Settings of \textsc{SafeTime}, grouped by role. Every setting is shared by all datasets, attacks, and
architectures.}
\label{tab:settings}
\footnotesize
\begin{tabular}{@{}ll@{}}
\toprule
Setting & Value \\
\midrule
\multicolumn{2}{@{}l}{\emph{Nominal level}} \\
level $\alpha$ & $10^{-3}$ \\
\midrule
\multicolumn{2}{@{}l}{\emph{Computational budget (larger values reduce sampling noise)}} \\
reference draws (tests; reversion) & $200$; $50$ \\
refinement rounds; trimming fits & $10$; $20$ \\
$k$-means restarts; random groups & $10$; $5$ \\
ridge penalty candidates (GCV) & $33$, log-spaced \\
co-extreme eigenvectors & $15$ \\
ridge penalty (co-extreme; residual) & $10^{-3}N$; $10^{-2}\times$ fit steps \\
\midrule
\multicolumn{2}{@{}l}{\emph{Minimum counts (for a statistic to be defined)}} \\
bursts per test & $3$ \\
channels for the cross-channel reference & $10$ (within-channel: $3$) \\
fit steps outside spans & $50$ \\
\midrule
\multicolumn{2}{@{}l}{\emph{Fixed design constants}} \\
graph links $K$; maximum communities & $10$; $15$ \\
confirmation cut on $z^{r}_c$ & $2.0$ \\
minimum group size & $2$ \\
residual regressors exclude & $c$, $\mathcal{D}$, $K$ neighbours \\
sentinel guard (training) & $3\times$ the $0.5$--$99.5\%$ range \\
\bottomrule
\end{tabular}}
\end{table}
\begin{table}[!t]\centering{\renewcommand{\arraystretch}{0.88}\scriptsize
\caption{Detection with each setting of Table~\ref{tab:settings} changed in turn, the others at their default (main poisoned series and clean series). All found: series with every attacked channel flagged; False clean: flags on the clean series. $^{\ast}$: random groups, random channels, bursts per test, empirical-null channels, and across-time minimum, each at either value.}
\label{tab:sensitivity}
\setlength{\tabcolsep}{2.5pt}
\begin{tabular}{llrrrr}
\toprule
Setting (\textsc{SafeTime}) & Value & Found$\uparrow$ & All found$\uparrow$ & False$\downarrow$ & False clean$\downarrow$  \\
\midrule
\textsc{SafeTime} & -- & 2{,}759 & 35 & 124 & 16 \\
\midrule
level $\alpha$ ($10^{-3}$) & $10^{-4}$ & 2{,}759 & 35 & 88 & 38 \\
 & $10^{-2}$ & 2{,}754 & 35 & 144 & 28 \\
null draws (tests) ($200$) & $100$ & 2{,}761 & 36 & 134 & 46 \\
 & $400$ & 2{,}761 & 35 & 118 & 54 \\
null draws (fallback) ($50$) & $25$ & 2{,}756 & 35 & 118 & 16 \\
 & $100$ & 2{,}762 & 36 & 133 & 16 \\
refinement rounds ($10$) & $5$ & 2{,}756 & 35 & 114 & 42 \\
 & $20$ & 2{,}761 & 35 & 124 & 16 \\
graph links $K$ ($10$) & $5$ & 2{,}760 & 36 & 241 & 15 \\
 & $20$ & 2{,}763 & 37 & 126 & 10 \\
max.\ communities ($15$) & $10$ & 2{,}763 & 37 & 123 & 18 \\
 & $25$ & 2{,}760 & 36 & 114 & 36 \\
$k$-means restarts ($10$) & $5$ & 2{,}761 & 35 & 123 & 21 \\
 & $20$ & 2{,}762 & 36 & 136 & 28 \\
five count settings$^{\ast}$ & either & 2{,}759 & 35 & 124 & 16 \\
co-extr.\ eigenvectors ($15$) & $8$ & 2{,}762 & 36 & 107 & 17 \\
 & $30$ & 2{,}760 & 36 & 109 & 20 \\
co-extr.\ ridge ($\times 10^{-3}N$) ($1$) & $0.5$ & 2{,}761 & 35 & 113 & 14 \\
 & $2$ & 2{,}762 & 36 & 105 & 16 \\
residual neighbours ($10$) & $5$ & 2{,}759 & 35 & 119 & 15 \\
 & $20$ & 2{,}759 & 35 & 159 & 12 \\
residual null draws ($200$) & $100$ & 2{,}760 & 36 & 147 & 42 \\
 & $400$ & 2{,}759 & 35 & 121 & 41 \\
residual ridge ($\times 10^{-2}$) ($1$) & $0.5$ & 2{,}759 & 35 & 122 & 17 \\
 & $2$ & 2{,}759 & 35 & 125 & 42 \\
confirmation cut ($2.0$) & $1.0$ & 2{,}759 & 35 & 139 & 45 \\
 & $4.0$ & 2{,}759 & 35 & 83 & 13 \\
\bottomrule
\end{tabular}}
\end{table}

\subsection{Empirical Results}

\subsubsection{Do the attacks implant? (RQ1)}
\label{sec:e-implant}

Measured by the implanted effect $I$, every attack family implants a backdoor in a victim trained from scratch.
\textsc{BackTime} reaches a mean implanted effect of $+0.644$ (Table~\ref{tab:main}), and \textsc{DEBA-MTS} and
\textsc{IFBA} reach $0.600$ to $0.998$ and $0.577$ to $0.961$ (Table~\ref{tab:deba}). The $17$ adaptive white-box poisons in Table~\ref{tab:adaptive}, $15$ full and $2$ channel-drop,
reach $0.542$ to $0.946$. \textsc{TDBA} implants less, $+0.185$ (Table~\ref{tab:main}). Table~\ref{tab:residual} shows that its trigger also acts on
clean-trained models ($+0.207$, and $+0.115$ for \textsc{BackTime}), which measures the
trigger response clean data alone teach.

\begin{table}[t]
\centering
\caption{\textsc{DEBA-MTS} and \textsc{IFBA}. Found, False: attacked and unattacked channels flagged. $I$: implanted effect. AMAE: attack MAE.}
\label{tab:deba}
\scriptsize
\setlength{\tabcolsep}{1.9pt}
\begin{tabular}{@{}lrrcccccc@{}}
\toprule
 & & & \multicolumn{3}{c}{Undefended} & \multicolumn{3}{c}{\textsc{SafeTime}} \\
\cmidrule(lr){4-6}\cmidrule(lr){7-9}
Dataset & Found$\uparrow$ & False$\downarrow$ & ASR$\downarrow$ & $I$$\downarrow$ & AMAE$\uparrow$ & ASR$\downarrow$ & $I$$\downarrow$ & AMAE$\uparrow$  \\
\midrule
\multicolumn{9}{@{}l}{\textsc{DEBA-MTS}~\cite{zhong2026deba}} \\
\textsc{PEMS03} & 107/107 & 0 & $0.925$ & $0.795$ & $10.9$ & $0.144$ & $-0.001$ & $37.2$ \\
\textsc{PEMS04} & 92/92 & 1 & $0.866$ & $0.600$ & $19.2$ & $0.255$ & $0.030$ & $41.6$ \\
\textsc{PEMS08} & 51/51 & 4 & $0.924$ & $0.779$ & $11.6$ & $0.150$ & $0.001$ & $36.9$ \\
\textsc{Weather} & 6/6 & 9 & $1.000$ & $0.989$ & $3.9$ & $0.015$ & $0.002$ & $85.4$ \\
\textsc{Electricity} & 96/96 & 12 & $0.999$ & $0.998$ & $168$ & $0.003$ & $0.001$ & $3436$ \\
\midrule
\multicolumn{9}{@{}l}{\textsc{IFBA}~\cite{li2026ifba}} \\
\textsc{PEMS03} & 108/108 & 1 & $0.915$ & $0.639$ & $14.1$ & $0.281$ & $0.034$ & $44.2$ \\
\textsc{PEMS04} & 92/93 & 0 & $0.876$ & $0.577$ & $19.5$ & $0.309$ & $0.051$ & $55.9$ \\
\textsc{PEMS08} & 51/51 & 0 & $0.919$ & $0.654$ & $14.2$ & $0.338$ & $0.031$ & $57.4$ \\
\textsc{Weather} & 7/7 & 5 & $0.999$ & $0.961$ & $6.4$ & $0.039$ & $-0.001$ & $92.8$ \\
\textsc{Electricity} & 97/97 & 2 & $0.925$ & $0.911$ & $1847$ & $0.098$ & $0.028$ & $5198$ \\
\bottomrule
\end{tabular}
\end{table}

\subsubsection{Does SafeTime remove the malicious supervision? (RQ2)}
\label{sec:e-main}

\begin{table*}[t]
\centering
\footnotesize
\caption{End-to-end results, mean over six architectures ($\pm$ standard deviation over training seeds). Undef.: undefended; No-trig.: undefended no-trigger rate; ST: \textsc{SafeTime}; TG: \textsc{TimeGuard}. Clean MAE: mean over the architectures and seeds; clean $\Delta$MAE: mean of each model's own change, so the two can differ in sign. Bold: better of ST and TG.}
\label{tab:main}
\setlength{\tabcolsep}{2.6pt}
\begin{tabular}{llcccccccccccc}
\toprule
 & & \multicolumn{4}{c}{ASR$\downarrow$} & \multicolumn{3}{c}{Attack MAE$\uparrow$} & \multicolumn{3}{c}{Clean MAE$\downarrow$} & \multicolumn{2}{c}{Clean $\Delta$MAE (\%)$\downarrow$}  \\
\cmidrule(lr){3-6}\cmidrule(lr){7-9}\cmidrule(lr){10-12}\cmidrule(lr){13-14}
Attack & Dataset & Undef. & No-trig. & ST & TG & Undef. & ST & TG & Undef. & ST & TG & ST & TG  \\
\midrule
\textsc{BackTime} & \textsc{PEMS03} & $0.787_{\pm 0.008}$ & $0.182$ & $\mathbf{0.162_{\pm 0.007}}$ & $0.201_{\pm 0.007}$ & 17.5 & \textbf{39.1} & 37.5 & 19.1 & \textbf{18.4} & 19.4 & $\mathbf{-3.6_{\pm 0.3}}$ & $+2.2_{\pm 0.4}$ \\
 & \textsc{PEMS04} & $0.707_{\pm 0.005}$ & $0.206$ & $\mathbf{0.198_{\pm 0.004}}$ & $0.240_{\pm 0.009}$ & 24.7 & \textbf{47.6} & 44.4 & 23.7 & \textbf{23.4} & 24.3 & $\mathbf{-1.3_{\pm 0.6}}$ & $+3.0_{\pm 1.0}$ \\
 & \textsc{PEMS08} & $0.746_{\pm 0.009}$ & $0.217$ & $\mathbf{0.196_{\pm 0.002}}$ & $0.244_{\pm 0.008}$ & 21.6 & \textbf{43.4} & 39.9 & 20.8 & \textbf{20.6} & 21.6 & $\mathbf{-0.5_{\pm 0.8}}$ & $+4.5_{\pm 1.7}$ \\
 & \textsc{Weather} & $0.880_{\pm 0.017}$ & $0.123$ & $\mathbf{0.112_{\pm 0.004}}$ & $0.692_{\pm 0.012}$ & 31.0 & \textbf{97.7} & 50.3 & 17.0 & \textbf{16.1} & 16.2 & $\mathbf{-4.7_{\pm 4.8}}$ & $+0.9_{\pm 2.5}$ \\
 & \textsc{Electricity} & $0.854_{\pm 0.019}$ & $0.028$ & $\mathbf{0.023_{\pm 0.001}}$ & $0.178_{\pm 0.093}$ & 1505 & \textbf{5068} & 3527 & 611 & \textbf{434} & 662 & $\mathbf{-30.2_{\pm 1.1}}$ & $+9.2_{\pm 2.0}$ \\
 & All & $0.795_{\pm 0.009}$ & $0.151$ & $\mathbf{0.138_{\pm 0.001}}$ & $0.311_{\pm 0.020}$ & -- & -- & -- & -- & -- & -- & $\mathbf{-8.0_{\pm 1.0}}$ & $+3.9_{\pm 0.7}$ \\
\midrule
\textsc{TDBA} & \textsc{PEMS03} & $0.614_{\pm 0.014}$ & $0.373$ & $\mathbf{0.317_{\pm 0.009}}$ & $0.595_{\pm 0.013}$ & 21.4 & \textbf{27.8} & 22.9 & 19.0 & \textbf{18.4} & 19.6 & $\mathbf{-2.9_{\pm 0.4}}$ & $+3.5_{\pm 0.3}$ \\
 & \textsc{PEMS04} & $0.533_{\pm 0.017}$ & $0.337$ & $\mathbf{0.290_{\pm 0.006}}$ & $0.534_{\pm 0.006}$ & 30.3 & \textbf{37.7} & 31.9 & 23.4 & \textbf{22.9} & 24.2 & $\mathbf{-1.8_{\pm 0.4}}$ & $+3.8_{\pm 0.0}$ \\
 & \textsc{PEMS08} & $0.546_{\pm 0.026}$ & $0.360$ & $\mathbf{0.304_{\pm 0.005}}$ & $0.575_{\pm 0.007}$ & 26.2 & \textbf{34.2} & 27.5 & 20.6 & \textbf{20.4} & 21.7 & $\mathbf{-0.6_{\pm 0.8}}$ & $+6.1_{\pm 0.6}$ \\
 & \textsc{Weather} & $0.645_{\pm 0.020}$ & $0.395$ & $\mathbf{0.352_{\pm 0.016}}$ & $0.688_{\pm 0.010}$ & 51.0 & \textbf{57.5} & 45.3 & 19.7 & \textbf{15.7} & 17.5 & $\mathbf{-12.8_{\pm 3.1}}$ & $+2.1_{\pm 5.7}$ \\
 & \textsc{Electricity} & $0.363_{\pm 0.007}$ & $0.311$ & $\mathbf{0.335_{\pm 0.006}}$ & $0.363_{\pm 0.005}$ & 1688 & 1800 & \textbf{1906} & 416 & \textbf{412} & 473 & $\mathbf{-1.5_{\pm 1.7}}$ & $+15.6_{\pm 1.4}$ \\
 & All & $0.540_{\pm 0.003}$ & $0.355$ & $\mathbf{0.320_{\pm 0.005}}$ & $0.551_{\pm 0.001}$ & -- & -- & -- & -- & -- & -- & $\mathbf{-3.9_{\pm 0.9}}$ & $+6.2_{\pm 1.0}$ \\
\bottomrule
\end{tabular}
\end{table*}

\begin{figure*}[t]
\centering
\includegraphics[width=\textwidth]{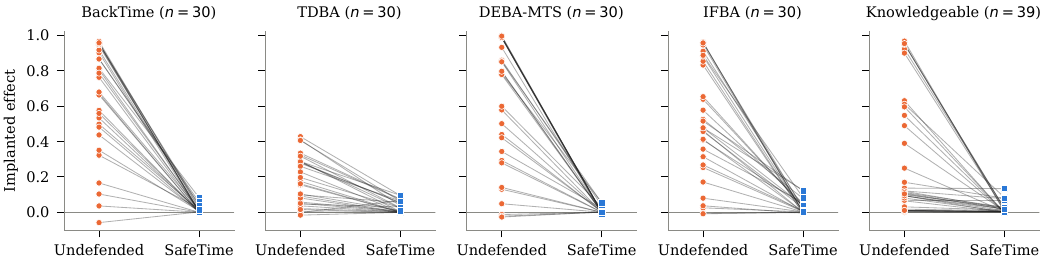}
\caption{Implanted effect undefended and after \textsc{SafeTime}; one line per poison and victim. \textsc{BackTime},
\textsc{TDBA}, \textsc{DEBA-MTS}, \textsc{IFBA}: every dataset on the six victims. Knowledgeable: ten lever combinations
(Section~\ref{sec:adaptive}) on \textsc{PEMS03}, \textsc{Electricity}, and \textsc{Weather}, and three on \textsc{PEMS04}
and \textsc{PEMS08}, on FEDformer.}
\label{fig:effect}
\end{figure*}

Table~\ref{tab:main} shows that \textsc{SafeTime} removes nearly all implanted effect. Against \textsc{BackTime}, the
mean implanted effect falls from $+0.644$ to $+0.013$ ($98.0\%$ removed) and the ASR from $0.795$ to $0.138$, below the
undefended no-trigger rate on every dataset and below the undefended ASR for every dataset, victim, and training seed;
against \textsc{TDBA}, the effect falls from $+0.185$ to $+0.030$ (ASR $0.540$ to $0.320$), and against \textsc{DEBA-MTS}
and \textsc{IFBA} to at most $0.030$ and $0.051$ (Table~\ref{tab:deba}), removals of at least $94.9\%$ and $91.1\%$. On
the other architectures (Table~\ref{tab:victims}, Appendix~\ref{app:arch}), no published-attack poison retains more than
$0.096$ (Figure~\ref{fig:effect}), except \textsc{IFBA} on iTransformer ($0.124$). Against \textsc{BackTime} and
\textsc{TDBA}, the residual effect is within $0.002$ and $0.008$ of perfect detection, which reconstructs every target of
the true attacked channels outside the true trigger and target steps ($+0.011$ and $+0.022$), and below that of the
clean-trained model ($+0.115$ and $+0.207$; Table~\ref{tab:residual}), so relabelling removes the trigger-conditional
deviation rather than masking it. Under the same information, \textsc{TimeGuard} leaves an ASR of $0.311$ and $0.551$
and an implanted effect of $0.181$ and $0.211$, and randomised smoothing, fine-tuning, and fine-pruning leave the ASR
near its undefended value (Table~\ref{tab:baselines}), as does the per-channel mask where it has an admissible setting
($0.896$ and $0.613$, against $0.893$ and $0.624$ undefended).

\begin{table}[t]
\centering
\footnotesize
\caption{Mean ASR and clean-MAE change ($\Delta$, \%) of \textsc{SafeTime} and the baselines. Bold: best without clean data. RS: randomised smoothing; FT: fine-tuning; FP: fine-pruning. $\dagger$: uses a clean slice of the training series. $^{a}$: mean over the cells with an admissible pruning ratio.}
\label{tab:baselines}
\setlength{\tabcolsep}{4.5pt}
\begin{tabular}{lcccc}
\toprule
 & \multicolumn{2}{c}{\textsc{BackTime}~\cite{lin2024backtime}} & \multicolumn{2}{c}{\textsc{TDBA}~\cite{liu2026tdba}}  \\
\cmidrule(lr){2-3}\cmidrule(lr){4-5}
Defence & ASR$\downarrow$ & $\Delta$$\downarrow$ & ASR$\downarrow$ & $\Delta$$\downarrow$  \\
\midrule
Undefended & 0.795 & -- & 0.540 & -- \\
\textsc{SafeTime} & \textbf{0.138} & $\mathbf{-8.0}$ & \textbf{0.320} & $\mathbf{-3.9}$ \\
\textsc{TimeGuard}~\cite{nguyen2026timeguard} & 0.311 & $+3.9$ & 0.551 & $+6.2$ \\
RS~\cite{cohen2019smoothing} & 0.776 & $+1.4$ & 0.537 & $+1.8$ \\
FT~\cite{liu2018finepruning} & 0.783 & $+2.9$ & 0.545 & $+3.8$ \\
FP~\cite{liu2018finepruning}$^{a}$ & 0.692 & $+2.5$ & 0.564 & $+3.8$ \\
\midrule
FT+clean$^\dagger$ & 0.552 & $-6.0$ & 0.482 & $-0.3$ \\
FP+clean$^{\dagger a}$ & 0.354 & $-5.0$ & 0.534 & $-0.2$ \\
\bottomrule
\end{tabular}
\end{table}

\begin{table}[t]
\centering
\footnotesize
\caption{Residual attack effect. Impl.: ASR minus the model's own no-trigger rate. $\Delta$MAE: clean-MAE change (\%).}
\label{tab:residual}
\setlength{\tabcolsep}{2.7pt}
\begin{tabular}{lcccccc}
\toprule
 & \multicolumn{3}{c}{\textsc{BackTime}~\cite{lin2024backtime}} & \multicolumn{3}{c}{\textsc{TDBA}~\cite{liu2026tdba}}  \\
\cmidrule(lr){2-4}\cmidrule(lr){5-7}
Model & ASR$\downarrow$ & Impl.$\downarrow$ & $\Delta$MAE$\downarrow$ & ASR$\downarrow$ & Impl.$\downarrow$ & $\Delta$MAE$\downarrow$  \\
\midrule
Undefended & 0.795 & $+0.644$ & -- & 0.540 & $+0.185$ & -- \\
\textsc{SafeTime} & 0.138 & $+0.013$ & $-8.0$ & 0.320 & $+0.030$ & $-3.9$ \\
Perfect detection & 0.132 & $+0.011$ & $-9.8$ & 0.307 & $+0.022$ & $-4.5$ \\
Clean-trained & 0.230 & $+0.115$ & $-10.7$ & 0.489 & $+0.207$ & $-6.0$ \\
\bottomrule
\end{tabular}
\end{table}

\subsubsection{Is detection attributable to target-side evidence? (RQ3)}
\label{sec:e-detection}
\label{sec:attribution}

\textsc{SafeTime} localises the attacked channels almost exactly. Its recall and precision are $0.998$ and $0.948$
against \textsc{BackTime} and $0.998$ and $0.986$ against \textsc{TDBA} (Table~\ref{tab:detection}), and $1.000$ and
$0.931$ against \textsc{DEBA-MTS} and $0.997$ and $0.978$ against \textsc{IFBA} (Table~\ref{tab:deba}).
Table~\ref{tab:attribution} removes the trigger, the shared lift, the target deviation, or the synchrony of the attacked
channels from \textsc{BackTime} and from a \textsc{PEMS03} poison whose targets add a shared lift to per-event shape
variation (Section~\ref{sec:adaptive}). The shared target deviation suffices: without the trigger, both poisons are
flagged on all $107$ attacked channels and still implant $0.652$ and $0.745$, whereas per-event shape variation alone,
the negative control, implants $0.085$ and is left unflagged. Under synchrony, the trigger also suffices: without a
shared target component, the poisons are flagged on all $107$ channels while implanting only $0.019$ to $0.025$. Without
synchrony, the lift poison is flagged on $88$ channels ($I = 0.048$), or $68$ when the trigger is also removed ($0.057$),
and \textsc{BackTime} on all $107$ ($0.083$).

\begin{table}[t]
\centering
\footnotesize
\caption{Channel-level detection on the poisoned series and channels flagged on the clean series. $C$: number of
channels. Recall: share of attacked channels flagged. Prec.: share of flagged channels that are attacked.}
\label{tab:detection}
\setlength{\tabcolsep}{4pt}
\begin{tabular}{lrccccr}
\toprule
 & & \multicolumn{2}{c}{\textsc{BackTime}~\cite{lin2024backtime}} & \multicolumn{2}{c}{\textsc{TDBA}~\cite{liu2026tdba}} & Clean  \\
\cmidrule(lr){3-4}\cmidrule(lr){5-6}
Dataset & $C$ & Recall$\uparrow$ & Prec.$\uparrow$ & Recall$\uparrow$ & Prec.$\uparrow$ & flags$\downarrow$  \\
\midrule
\textsc{PEMS03} & 358 & 1.000 & 0.991 & 1.000 & 0.991 & 2 \\
\textsc{PEMS04} & 307 & 1.000 & 1.000 & 1.000 & 1.000 & 0 \\
\textsc{PEMS08} & 170 & 1.000 & 0.930 & 1.000 & 0.981 & 3 \\
\textsc{Weather} & 21 & 1.000 & 0.750 & 1.000 & 0.750 & 8 \\
\textsc{Electricity} & 321 & 0.991 & 0.885 & 0.990 & 0.979 & 3 \\
\midrule
All & & 0.998 & 0.948 & 0.998 & 0.986 & 16 \\
\bottomrule
\end{tabular}
\end{table}

\begin{table}[t]
\centering
\caption{Evidence removed from \textsc{PEMS03} poisons (first row of each block: complete). Top: shared lift plus per-event shape variation; bottom: \textsc{BackTime}. Sync.: synchronised events. $I$: undefended implanted effect. $\dagger$: negative control.}
\label{tab:attribution}
\footnotesize
\setlength{\tabcolsep}{4pt}
\begin{tabular}{lllrrc}
\toprule
Trigger & Target deviation & Sync. & Found & False & $I$ \\
\midrule
yes & lift + varied shape & yes & 107/107 & 1 & $0.748$ \\
no & lift + varied shape & yes & 107/107 & 1 & $0.652$ \\
yes & none & yes & 107/107 & 1 & $0.019$ \\
yes & varied shape only & yes & 107/107 & 1 & $0.025$ \\
no & varied shape only$^\dagger$ & yes & 0/107 & 1 & $0.085$ \\
yes & lift + varied shape & no & 88/107 & 1 & $0.048$ \\
no & lift + varied shape & no & 68/107 & 1 & $0.057$ \\
\midrule
yes & published target & yes & 107/107 & 1 & $0.806$ \\
no & published target & yes & 107/107 & 1 & $0.745$ \\
yes & none & yes & 107/107 & 1 & $0.023$ \\
yes & published target & no & 107/107 & 1 & $0.083$ \\
\bottomrule
\end{tabular}
\end{table}

\subsubsection{Which components are needed? (RQ4)}
\label{sec:e-ablation}

\begin{table}[t]
\centering
\footnotesize
\caption{Detector ablation on the poisoned series of the main evaluation ($2{,}765$ attacked channels) and the clean series; each row removes one component. Found: attacked channels flagged. False: unattacked channels flagged.}
\label{tab:ablation}
\setlength{\tabcolsep}{2.3pt}
\begin{tabular}{lrrr}
\toprule
 & Attacked & False & False \\
Variant & found$\uparrow$ & (poisoned)$\downarrow$ & (clean)$\downarrow$  \\
\midrule
\textsc{SafeTime} & 2{,}759 & 124 & 16 \\
\midrule
No cross-channel test & 915 & 9 & 11 \\
No co-extreme groups & 2{,}755 & 132 & 7 \\
No within-channel test & 2{,}759 & 76 & 37 \\
No residual stage & 2{,}755 & 173 & 47 \\
No confirmation & 2{,}759 & 184 & 51 \\
No expansion & 2{,}755 & 113 & 12 \\
No group confirmation & 2{,}758 & 112 & 16 \\
All group members kept & 2{,}759 & 128 & 16 \\
No refinement fallback & 2{,}674 & 76 & 31 \\
No refinement & 2{,}762 & 253 & 63 \\
No cross-channel reference & 2{,}765 & 965 & 213 \\
Level view only & 2{,}621 & 62 & 20 \\
Change view only & 2{,}751 & 82 & 24 \\
No leave-one-out & 2{,}762 & 121 & 45 \\
Burst sample kept & 2{,}759 & 206 & 39 \\
Global burst scoring & 2{,}763 & 394 & 42 \\
\bottomrule
\end{tabular}
\end{table}

Every component contributes recall or precision. Each row of Table~\ref{tab:ablation} removes one component and reruns
detection on the main and clean series. \textsc{SafeTime} finds $2{,}759$ of the $2{,}765$ attacked channels. The cross-channel test supplies most of the recall:
without it, $915$ are found. The co-extreme groups and expansion each add $4$ channels. The within-channel test, which
detects attacks on a single channel (Appendix~\ref{app:attackers}), changes no detection here: without it, false flags
fall from $124$ to $76$ on the poisoned series but rise from $16$ to $37$ on the clean series. Phase matching (scoring
bursts against the same phase rather than globally), the cross-channel reference, and residual confirmation supply
precision. Without the residual stage, false flags rise from $124$ to $173$ on the poisoned and from $16$ to $47$ on the
clean series. Without the cross-channel reference they reach $965$ and $213$, and with global burst scoring $394$ and
$42$.

\subsubsection{Adaptive white-box evaluation (RQ5)}
\label{sec:adaptive}~ 

{\textbf{Attacker}.} The attacker knows \textsc{SafeTime}'s algorithm, settings, and the defender's exact
series, runs \textsc{SafeTime} locally, and seeks a backdoor that survives relabelling by diversifying the planted
targets while keeping the shared deviation that, by Proposition~1, any learnable backdoor requires. It screens
candidates with \textsc{SafeTime}'s detection only; victims are retrained for the final poisons.

{\textbf{Variants and budget}.} Starting from the \textsc{BackTime} poison on \textsc{PEMS03} ($107$
attacked channels), \textsc{Electricity} ($96$), \textsc{PEMS04} ($92$), \textsc{PEMS08} ($51$), and \textsc{Weather}
($6$), the attacker rewrites only the targets. Fixed-goal diversification adds to each copy of the published target an
independent segment of the channel's short-scale variation, scaled by $\rho$. Outcome-goal diversification keeps only
the published target's mean lift, times one of five mean-one profiles (flat, rising, falling, hump, late), plus a
natural segment, each copy scored on its own goal. Detector-guided variants redraw every flagged channel's copies with
$\rho$ raised by $0.25$, for six rounds with $\rho \leq 2$ ($70$ candidate poisons over the five datasets); channel-drop
variants then revert the channels still flagged to the clean series, and controls desynchronise the events across
channels. Every final poison is measured on an undefended victim.

\begin{table*}[t]
\centering
\caption{Adaptive white-box attacker. Goal: fixed (published target), outcome (shared lift, any shape), or direction (shared zero-mean rising profile, no lift). Feedback: redraws of flagged channels. Channels: all kept, or flagged channels dropped. Bottom: controls without cross-channel timing coordination. Found, False: attacked and unattacked channels flagged. $I$: implanted effect. Removal: $1 - I_{\text{after}}/I_{\text{before}}$ (\%). $\Delta$MAE: clean-MAE change (\%).}
\label{tab:adaptive}
\scriptsize
\setlength{\tabcolsep}{2.9pt}
\begin{tabular}{llllrrccccccrr}
\toprule
 & & & & & & \multicolumn{2}{c}{$I$} & \multicolumn{2}{c}{Attack MAE} & \multicolumn{2}{c}{Clean MAE$\downarrow$} & & \\
\cmidrule(lr){7-8}\cmidrule(lr){9-10}\cmidrule(lr){11-12}
Dataset & Goal & Feedback & Channels / timing & Found & False & Undef. & \textsc{SafeTime} & Undef. & \textsc{SafeTime} & Undef. & \textsc{SafeTime} & Removal & $\Delta$MAE \\
\midrule
\textsc{PEMS03} & fixed & yes & all & 107/107 & 1 & $0.798$ & $0.011$ & $11.7$ & $38.2$ & $16.2$ & $16.0$ & $98.6$ & $-0.9$ \\
 & outcome & no & all & 107/107 & 1 & $0.748$ & $0.005$ & $12.0$ & $38.0$ & $16.3$ & $16.0$ & $99.4$ & $-1.8$ \\
 & outcome & yes & all & 107/107 & 1 & $0.725$ & $0.003$ & $12.2$ & $36.7$ & $16.3$ & $16.0$ & $99.6$ & $-1.4$ \\
 & direction & no & all & 107/107 & 1 & $0.894$ & $0.017$ & $14.4$ & $37.9$ & $15.9$ & $16.0$ & $98.1$ & $+0.2$ \\
 & direction & yes & all & 107/107 & 1 & $0.870$ & $0.018$ & $15.9$ & $38.0$ & $16.0$ & $16.0$ & $98.0$ & $-0.3$ \\
\midrule
\textsc{PEMS04} & fixed & yes & all & 92/92 & 0 & $0.662$ & $0.014$ & $18.3$ & $43.7$ & $21.1$ & $21.1$ & $97.9$ & $-0.3$ \\
 & outcome & no & all & 92/92 & 0 & $0.614$ & $0.006$ & $18.4$ & $43.5$ & $21.2$ & $21.2$ & $99.1$ & $+0.2$ \\
 & outcome & yes & all & 92/92 & 0 & $0.606$ & $0.004$ & $19.0$ & $43.5$ & $21.1$ & $21.2$ & $99.3$ & $+0.3$ \\
\midrule
\textsc{PEMS08} & fixed & yes & all & 51/51 & 6 & $0.756$ & $-0.000$ & $13.2$ & $40.7$ & $17.6$ & $17.8$ & $100.0$ & $+1.3$ \\
 & outcome & no & all & 51/51 & 1 & $0.695$ & $-0.006$ & $12.7$ & $39.6$ & $17.8$ & $17.9$ & $100.9$ & $+1.0$ \\
 & outcome & yes & all & 51/51 & 1 & $0.699$ & $-0.008$ & $13.2$ & $39.3$ & $17.7$ & $17.9$ & $101.2$ & $+1.1$ \\
\midrule
\textsc{Weather} & fixed & yes & all & 6/6 & 2 & $0.943$ & $0.002$ & $12.3$ & $90.8$ & $10.4$ & $10.3$ & $99.8$ & $-1.7$ \\
 & outcome & no & all & 6/6 & 2 & $0.932$ & $0.005$ & $13.0$ & $93.0$ & $10.5$ & $9.7$ & $99.5$ & $-7.3$ \\
 & outcome & yes & all & 6/6 & 2 & $0.929$ & $0.003$ & $16.2$ & $96.0$ & $9.9$ & $9.2$ & $99.7$ & $-6.8$ \\
\midrule
\textsc{Electricity} & fixed & yes & all & 95/96 & 9 & $0.936$ & $0.008$ & $944$ & $4229$ & $529$ & $397$ & $99.1$ & $-24.9$ \\
 & outcome & no & all & 95/96 & 7 & $0.946$ & $0.015$ & $922$ & $4212$ & $471$ & $410$ & $98.4$ & $-12.9$ \\
 & outcome & yes & all & 95/96 & 6 & $0.938$ & $0.009$ & $996$ & $4259$ & $492$ & $394$ & $99.0$ & $-20.0$ \\
 & direction & no & all & 95/96 & 8 & $0.937$ & $0.007$ & $1267$ & $4170$ & $435$ & $399$ & $99.2$ & $-8.3$ \\
 & direction & yes & all & 95/96 & 7 & $0.932$ & $0.007$ & $1287$ & $4181$ & $434$ & $395$ & $99.2$ & $-8.9$ \\
 & fixed & yes & drop & 0/1 & 3 & $0.542$ & $0.495$ & $96772$ & $108152$ & $401$ & $402$ & $8.7$ & $+0.2$ \\
 & outcome & yes & drop & 0/1 & 2 & $0.565$ & $0.559$ & $93823$ & $96523$ & $398$ & $397$ & $1.0$ & $-0.3$ \\
\midrule
\textsc{PEMS03} & outcome & no & per-channel offset, shared trig. & 88/107 & 1 & $0.048$ & $0.037$ & $33.2$ & $34.2$ & $16.4$ & $16.3$ & -- & $-0.5$ \\
 & outcome & no & random per event, varied trig. & 51/107 & 1 & $0.045$ & $0.044$ & $34.4$ & $34.9$ & $16.4$ & $16.4$ & -- & $0.0$ \\
 & fixed & no & random per event, varied trig. & 60/107 & 1 & $0.039$ & $0.039$ & $35.5$ & $36.4$ & $16.5$ & $16.3$ & -- & $-0.6$ \\
\bottomrule
\end{tabular}
\end{table*}

{\textbf{Outcome}.} \textsc{SafeTime} flags every attacked channel in every round on four datasets and $95$ of
$96$ on \textsc{Electricity} (final round in Table~\ref{tab:adaptive}). The full fixed- and outcome-goal poisons implant $0.606$ to $0.946$, and
retraining on \textsc{SafeTime}'s labels leaves at most $0.015$, a removal of at least $97.9\%$
(Table~\ref{tab:adaptive}). Detector feedback gives the attacker no advantage: guided poisons keep at most $0.009$,
against $0.015$ unguided, and at most $0.039$ on any of the six architectures in Table~\ref{tab:victims}. Without
timing coordination across channels, the controls implant only $0.039$ to $0.048$. The one successful adaptive attack uses the \textsc{Electricity} channel that is never flagged:
the channel-drop variants revert the other $95$ channels and keep a single-channel backdoor, which implants $0.542$ and
$0.565$ and keeps $0.495$ and $0.559$ after \textsc{SafeTime}, at the cost of $95$ of the attacker's $96$ channels. On
the other four datasets, no attacked channel survives.

{\textbf{Shared direction without lift}.} Keeping only a shared direction, each planted target adds to the
clean value one of five zero-mean rising profiles, scaled to the channel's variability, plus a natural segment, with
and without feedback. On \textsc{PEMS03} and \textsc{Electricity}, these poisons implant $0.870$ to $0.945$ on FEDformer
and iTransformer; \textsc{SafeTime} flags $107/107$ and $95/96$ attacked channels and leaves $0.007$ to $0.042$
(Tables~\ref{tab:adaptive} and~\ref{tab:victims}).

\begin{table*}[!t]\centering{
\caption{Sign-changing deviation: each target follows the channel's pre-trigger direction, so the planted deviation averages to about zero. Feedback: redraws of flagged channels. $I$ and attack MAE are measured on this context-dependent goal. Removal: $1 - I_{\text{after}}/I_{\text{before}}$ (\%).}
\label{tab:signflip}
\footnotesize
\setlength{\tabcolsep}{3pt}
\begin{tabular}{@{}lllrrrrrrrr@{}}
\toprule
 & & & & & \multicolumn{2}{c}{$I$$\downarrow$} & \multicolumn{2}{c}{Attack MAE$\uparrow$} & &  \\
\cmidrule(lr){6-7}\cmidrule(lr){8-9}
Dataset & Feedback & Victim & Found$\uparrow$ & False$\downarrow$ & Undef. & \textsc{SafeTime} & Undef. & \textsc{SafeTime} & Removal$\uparrow$ & $\Delta$MAE$\downarrow$  \\
\midrule
\textsc{PEMS03} & no & FEDformer & 107/107 & 1 & $0.064$ & $0.016$ & 29.2 & 32.4 & $74.6$ & $-0.7$ \\
 &  & iTransformer & 107/107 & 1 & $0.577$ & $0.045$ & 14.2 & 31.2 & $92.1$ & $+0.1$ \\
 & yes & FEDformer & 107/107 & 1 & $0.050$ & $0.016$ & 29.8 & 32.4 & $67.2$ & $+0.2$ \\
 &  & iTransformer & 107/107 & 1 & $0.552$ & $0.049$ & 15.2 & 31.2 & $91.2$ & $+0.2$ \\
\midrule
\textsc{Electricity} & no & FEDformer & 95/96 & 2 & $0.688$ & $0.010$ & 2112 & 4170 & $98.5$ & $-10.1$ \\
 &  & iTransformer & 95/96 & 2 & $0.703$ & $0.025$ & 2483 & 4312 & $96.5$ & $-4.1$ \\
 & yes & FEDformer & 95/96 & 4 & $0.681$ & $0.011$ & 2126 & 4216 & $98.4$ & $-11.7$ \\
 &  & iTransformer & 95/96 & 4 & $0.701$ & $0.024$ & 2534 & 4501 & $96.6$ & $+3.4$ \\
\bottomrule
\end{tabular}}
\end{table*}

{\textbf{Sign-changing deviation}.} For the case not covered by the averaging argument of
Section~\ref{sec:i-proposition}, the same attacker pushes each outcome-goal target in the direction the channel moved before the
trigger, split at the channel's median so that the deviation averages to about zero.
\textsc{SafeTime} flags $107/107$ and $95/96$ attacked channels with and without feedback. Where the poison implants,
with effects of $0.552$ to $0.703$, retraining leaves $0.010$ to $0.049$ (Table~\ref{tab:signflip}); on FEDformer, the
\textsc{PEMS03} poison implants only $0.050$ and $0.064$.

{\textbf{Knowledgeable lever combinations}.} A knowledgeable attacker combines levers aimed at
\textsc{SafeTime}'s assumptions: fewer channels or events; smaller, shorter, smooth, or diversified triggers; and
camouflage, timing events to moments when the attacked channels already move together. Of $30$ combinations on
\textsc{PEMS03}, \textsc{Electricity}, and \textsc{Weather} ($17$ in Table~\ref{tab:knowledgeable}; the other $13$, and three run
on \textsc{PEMS04} and \textsc{PEMS08}, in Table~\ref{tab:kn-rest} where they implant), every multichannel combination of
Table~\ref{tab:knowledgeable} is flagged on all attacked channels or all but one, with a defended ASR of at most $0.136$
and a higher attack MAE. On \textsc{Weather}, the ASR of C R S falls from $0.986$ to $0.036$, against $0.483$ for
\textsc{TimeGuard}, and stays at or below $0.04$ on every victim in Table~\ref{tab:transfer}. One configuration evades
detection while still implanting: a single \textsc{PEMS03} channel with few events and a small trigger ($0.135$;
Table~\ref{tab:knowledgeable}). In Table~\ref{tab:kn-rest}, the \textsc{PEMS03} corner combination also evades detection
but barely implants ($0.067$), and the \textsc{PEMS04} C R combination, flagged on only $11$ of its $31$ channels, still
has its ASR lowered from $0.297$ to $0.253$.

\begin{table*}[t]
\centering
\caption{Knowledgeable attacker. Levers: C, $10\%$ of channels attacked; R, $1\%$ poisoning rate; N, four-fold trigger-magnitude penalty; L, two-step trigger; S, smooth trigger; E, diversified triggers; T, camouflage timing; O, one attacked channel. Found: attacked channels flagged. Undef., No-trig.: undefended ASR and no-trigger rate; Impl.: their difference; ST: \textsc{SafeTime}; TG: \textsc{TimeGuard}; Clean-tr.: clean-trained model. Bold: better of ST and TG.}
\label{tab:knowledgeable}
\footnotesize\setlength{\tabcolsep}{3pt}
\begin{tabular}{llrcccccccccccc}
\toprule
 & & & \multicolumn{6}{c}{ASR$\downarrow$} & \multicolumn{3}{c}{Attack MAE$\uparrow$} & \multicolumn{3}{c}{Clean MAE$\downarrow$} \\
\cmidrule(lr){4-9}\cmidrule(lr){10-12}\cmidrule(lr){13-15}
Dataset & Levers & Found$\uparrow$ & Undef. & No-trig. & Impl. & ST & TG & Clean-tr. & Undef. & ST & TG & Undef. & ST & TG  \\
\midrule
\textsc{PEMS03} & C R & 35/36 & 0.244 & 0.131 & 0.113 & \textbf{0.136} & 0.175 & 0.183 & 29.1 & \textbf{36.9} & 31.4 & 15.8 & \textbf{15.7} & 16.8 \\
 & C R N & 35/36 & 0.230 & 0.131 & 0.099 & \textbf{0.131} & 0.164 & 0.185 & 29.7 & \textbf{37.2} & 32.2 & 15.9 & \textbf{15.8} & 16.8 \\
 & C R S & 36/36 & 0.615 & 0.134 & 0.481 & \textbf{0.131} & 0.253 & 0.321 & 22.1 & \textbf{38.5} & 27.8 & 15.7 & \textbf{15.8} & 16.8 \\
 & C R E & 36/36 & 0.425 & 0.133 & 0.292 & \textbf{0.121} & 0.181 & 0.185 & 24.6 & \textbf{39.2} & 31.8 & 15.8 & \textbf{15.7} & 16.8 \\
 & C R T & 36/36 & 0.709 & 0.123 & 0.586 & \textbf{0.116} & 0.166 & 0.178 & 20.9 & \textbf{39.7} & 31.9 & 15.7 & \textbf{15.7} & 16.7 \\
 & C R N E T & 36/36 & 0.733 & 0.125 & 0.608 & \textbf{0.118} & 0.169 & 0.177 & 20.0 & \textbf{39.0} & 31.6 & 15.8 & \textbf{15.7} & 16.7 \\
 & O R N & 0/1 & 0.232 & 0.097 & 0.135 & 0.228 & \textbf{0.164} & 0.191 & 25.8 & 25.6 & \textbf{26.0} & 15.7 & \textbf{15.7} & 16.8 \\
 & O E T & 1/1 & 0.296 & 0.095 & 0.202 & 0.142 & \textbf{0.130} & 0.196 & 23.9 & \textbf{29.0} & 28.6 & 15.7 & \textbf{15.8} & 16.7 \\
\midrule
\textsc{Electricity} & C R & 32/32 & 0.893 & 0.004 & 0.890 & \textbf{0.017} & 0.019 & 0.055 & 1701 & \textbf{4589} & 4148 & 302 & \textbf{301} & 405 \\
 & C R N & 31/32 & 0.954 & 0.003 & 0.951 & 0.027 & \textbf{0.024} & 0.097 & 2414 & \textbf{5840} & 5364 & 312 & \textbf{312} & 390 \\
 & C R N L & 31/32 & 0.890 & 0.005 & 0.886 & \textbf{0.019} & \textbf{0.019} & 0.011 & 3406 & 5915 & \textbf{6180} & 313 & \textbf{313} & 385 \\
 & C R S & 31/32 & 0.613 & 0.004 & 0.609 & 0.084 & \textbf{0.030} & 0.024 & 3703 & 5415 & \textbf{5602} & 311 & \textbf{314} & 394 \\
 & C R E & 31/32 & 0.954 & 0.003 & 0.951 & 0.026 & \textbf{0.024} & 0.099 & 2442 & \textbf{5765} & 5330 & 309 & \textbf{310} & 389 \\
 & C R T & 32/32 & 0.015 & 0.004 & 0.011 & 0.008 & \textbf{0.007} & 0.011 & 2711 & \textbf{3503} & 3168 & 289 & \textbf{290} & 412 \\
 & C R N E T & 31/32 & 0.356 & 0.003 & 0.353 & \textbf{0.008} & 0.014 & 0.019 & 2168 & \textbf{3748} & 3289 & 289 & \textbf{289} & 411 \\
 & O E T & 1/1 & 0.456 & 0.000 & 0.456 & \textbf{0.000} & 0.001 & 0.001 & 1877 & \textbf{3234} & 2821 & 290 & \textbf{293} & 417 \\
\midrule
\textsc{Weather} & C R S & 2/2 & 0.986 & 0.032 & 0.954 & \textbf{0.036} & 0.483 & 0.045 & 34.2 & \textbf{83.3} & 49.2 & 9.6 & \textbf{9.9} & 10.8 \\
\bottomrule
\end{tabular}
\end{table*}

{\textbf{Stronger fixed attackers}.} Stealthier \textsc{BackTime} settings, a single poisoned channel, several
simultaneous attackers, random variation of every planted copy, and half of the channels attacked are evaluated in
Appendix~\ref{app:attackers}.

\begin{table}[!t]\centering{
\caption{Attacks on the reconstruction (\textsc{BackTime}; Section~\ref{sec:adaptive}). \textsc{SafeTime}'s ASR on the base poisons is $0.069$; all $216$ decoys are flagged. $\Delta$: median clean-MAE change (\%).}
\label{tab:supp-recon}
\scriptsize
\setlength{\tabcolsep}{3pt}
\begin{tabular}{@{}lccccc@{}}
\toprule
Attack & Undef. & ST & TG & Found & $\Delta$ \\
\midrule
Echo, $\epsilon = 0.25$ & $0.962$ & $0.106$ & $0.276$ & $626/627$ & $+0.4$ \\
Echo, $\epsilon = 0.5$ & $0.964$ & $0.105$ & $0.294$ & $626/627$ & $+1.4$ \\
Echo, $\epsilon = 1$ & $0.961$ & $0.128$ & $0.390$ & $626/627$ & $+5.9$ \\
False-flag decoys & $0.957$ & $0.071$ & $0.252$ & $626/627$ & $+1.6$ \\
\bottomrule
\end{tabular}}
\end{table}

{\textbf{Attacks on the reconstruction}.} An echo of at most one standard deviation on the unflagged channels
that most influence the reconstruction, solved to rebuild the target, raises \textsc{SafeTime}'s mean ASR from $0.069$
to at most $0.128$, against $0.276$ to $0.390$ for \textsc{TimeGuard}. False-flag decoys planted on
$10\%$ of the channels are all flagged, while $626$ of $627$ attacked channels are still found. Table~\ref{tab:supp-recon}
gives each setting.

\subsubsection{Is forecasting utility preserved? (RQ6)}
\label{sec:e-utility}

\begin{table}[!t]\centering{\scriptsize
\caption{Cost of false alarms on clean data: change in clean MAE (\%; lower is better) of each architecture trained on the clean series after \textsc{SafeTime}, relative to the same architecture trained on the unmodified clean series.}
\label{tab:cleancost}
\setlength{\tabcolsep}{2.5pt}
\begin{tabular}{lccccc}
\toprule
Model & \textsc{PEMS03} & \textsc{PEMS04} & \textsc{PEMS08} & \textsc{Weather} & \textsc{Electricity} \\
\midrule
FEDformer & $+1.64$ & $-0.28$ & $+1.40$ & $-2.60$ & $-0.46$ \\
Autoformer & $-0.04$ & $+1.24$ & $-0.51$ & $+0.26$ & $-3.11$ \\
TimesNet & $-0.70$ & $+0.00$ & $+0.04$ & $-13.02$ & $+0.01$ \\
Informer & $+0.46$ & $+0.54$ & $-2.50$ & $+37.30$ & $+0.07$ \\
PatchTST & $+0.10$ & $+0.00$ & $+0.01$ & $+2.19$ & $-0.10$ \\
iTransformer & $-0.35$ & $-0.19$ & $-1.71$ & $+2.28$ & $+2.84$ \\
\midrule
Mean & $+0.19$ & $+0.22$ & $-0.54$ & $+4.40$ & $-0.13$ \\
\bottomrule
\end{tabular}}
\end{table}

In the main evaluation, \textsc{SafeTime} lowers the average clean MAE by $8.0\%$ (\textsc{BackTime}) and $3.9\%$
(\textsc{TDBA}), most on \textsc{Electricity} ($-30.2\%$), where poisoning itself raises the clean error: the mean clean
MAE of the \textsc{BackTime} models is $611$ undefended, $434$ with \textsc{SafeTime}, $393$ with perfect detection, and
$369$ when trained on the clean series. It rises by more than $5\%$ on no dataset and victim (Table~\ref{tab:cells}), whereas
\textsc{TimeGuard} raises the mean by $3.9\%$ and $6.2\%$ (Table~\ref{tab:baselines}). It changes
by $-24.9\%$ to $+1.3\%$ on the full adaptive poisons of Table~\ref{tab:adaptive}. It rises on \textsc{Weather}
under \textsc{DEBA-MTS} (from $10.12$ to $30.34$, with $9$ false flags) and by $4.7\%$ to $31.6\%$ on four datasets
under \textsc{IFBA}, on FEDformer. On the clean series, the mean change per dataset is at most $+0.22\%$, except on \textsc{Weather} ($+4.40\%$; Table~\ref{tab:cleancost}).

\subsubsection{Is SafeTime practical? (RQ7)}
\label{sec:e-cost}

\textsc{SafeTime} runs once per training series on one CPU core of an Intel Xeon Cascade Lake node, with
single-threaded numerical libraries and a $5$\,GB memory allocation, independently of the architecture. On the main
series, detection and relabelling took a median of $796$, $380$, $231$, $75$, and $785$\,s on \textsc{PEMS03},
\textsc{PEMS04}, \textsc{PEMS08}, \textsc{Weather}, and \textsc{Electricity}. Every victim is then trained as an
undefended model, and inference is unchanged. \textsc{TimeGuard} instead acts during training and adds GPU time
(NVIDIA A100 or L40S) to every model it protects, a median of $837$, $535$, $382$, $301$, and $924$\,s per model on the
same series. The cost therefore depends on how many victims a series trains. With a single victim, \textsc{SafeTime}'s
one-off cost is $1.05$ to $4.0$ times lower than \textsc{TimeGuard}'s overhead; with all six architectures,
\textsc{TimeGuard} adds $1{,}805$ to $5{,}542$\,s per series, $6.3$ to $24$ times \textsc{SafeTime}'s.
\section{Discussion and Limitations}
\label{sec:discussion}
Relabelling cost depends on false flags and on how well the remaining channels predict a flagged one. On series with
few channels, false flags raise the clean error without restoring the backdoor: \textsc{SafeTime} flags $8$ of the $21$
\textsc{Weather} channels on its clean series, the largest clean-data cost ($+4.40\%$, Table~\ref{tab:cleancost}).
Relabelling only the detected spans may lower this cost. Residual effects on architectures such as iTransformer, which
respond to the triggers even when trained on clean data, reflect the forecaster's own robustness. An attacker that
reverts every flagged channel to the clean series keeps any channel that \textsc{SafeTime} never flags, at the cost of
the others (Table~\ref{tab:adaptive}).

\section{Related Work}
\label{sec:related}

{\textbf{Backdoor attacks and defences}.} Backdoors were first demonstrated on image
classifiers~\cite{gu2019badnets}, then extended to other models~\cite{liu2018trojaning} and to time-series
classification~\cite{ding2022tsbackdoor,jiang2023tsbackdoor}. Forecasting
attacks~\cite{lin2024backtime,liu2026tdba,zhong2026deba,li2026ifba} differ in the trigger, but all write the target
supervision that \textsc{SafeTime} analyses. Model-side defences need resources our defender lacks: clean
data~\cite{wang2019neuralcleanse,liu2018finepruning,zhu2023selectiveamnesia}, a classifier's predictions or output
classes~\cite{gao2019strip,wang2024mmbd}, or, for regression, white-box access and benign data~\cite{du2024drmguard};
the failure analysis of Zhang et al.\ concerns image classifiers~\cite{zhang2024orthogonality}. Data-side defences score
examples through a model trained on them, by spectral signatures~\cite{tran2018spectral,hayase2021spectre}, activation
clustering~\cite{chen2018activation}, or learning speed~\cite{li2021abl,jiang2024e2abl}, or test feature--label
co-occurrence in text~\cite{he2023spurious}. The closest work, \textsc{TimeGuard}~\cite{nguyen2026timeguard}, counters
the dilution of poisoned signals across channels with channel-wise pools of reliable data; \textsc{SafeTime} instead tests
target supervision before training, and relabels rather than discards.

{\textbf{Repeated patterns and robust learning}.} Motif discovery finds repeated subsequences, including
multichannel ones~\cite{yeh2017mstamp}, with significance assessed against a Markov-chain model that does not represent
the daily cycle~\cite{castro2012motifs}. Nearest-neighbour discord detectors can miss an anomaly that occurs twice, which
DAMP~\cite{lu2022damp} is designed to avoid; a backdoor recurs by design. Cyber-physical detectors predict each sensor
from the others~\cite{ghafouri2018adversarial}, locate attacked channels by low-rank
decomposition~\cite{gao2016unobservable}, or flag deviations from learned normal
behaviour~\cite{audibert2020usad,deng2021gdn}, all on online measurements, and can be bypassed by false-data injections
crafted with knowledge of the system configuration~\cite{liu2009fdi}. Trimmed
regression~\cite{jagielski2018manipulating,jiao2022resilient}, robust estimators~\cite{rousseeuw1984lms}, and outlier
removal with worst-case bounds~\cite{steinhardt2017certified} discard points that fit badly or look atypical, whereas a
backdoor target is designed to be learned. Randomised smoothing~\cite{cohen2019smoothing}, also used in
forecasting~\cite{yoon2022robustforecast,liu2023robustmts}, addresses test-time perturbations.

\section{Conclusion}
\label{sec:conclusion}

\textsc{SafeTime} shifts forecasting-backdoor defence from trigger inspection to target-side analysis. It tests the
training targets for the trigger-conditional deviation that a forecaster must learn, against phase-matched natural
dynamics, and relabels those that carry it before training. Across attacks, datasets, architectures, and adaptive
white-box target diversification, it removes most implanted effect. Target-side supervision is therefore a practical
defensive surface for forecasters trained on untrusted histories. 

\bibliographystyle{IEEEtran}
\bibliography{refs}

\appendices
\raggedbottom
\setlength{\floatsep}{4pt plus 2pt minus 1pt}

\newcommand{\appinner}{\renewcommand{\arraystretch}{0.88}%
  \renewenvironment{table*}[1][]{\par\vspace{6pt}}{\par}%
  \renewenvironment{table}[1][]{\par\vspace{6pt}}{\par}}
\newcommand{\appwide}[1]{\begin{table*}[!t]\centering{\appinner\vspace{-8pt}#1}\end{table*}}
\newcommand{\appcol}[1]{\begin{table}[!t]\centering{\appinner\vspace{-8pt}#1}\end{table}}
\newcommand{\apphere}[1]{\begin{table}[H]\centering{\appinner\vspace{-8pt}#1}\end{table}}

\makeatletter\newcommand\captionoftable{\def\@captype{table}\caption}\newcommand\captionoffigure{\def\@captype{figure}\caption}\makeatother

\setcounter{table}{16}
\begin{table*}[!t]\centering{\renewcommand{\arraystretch}{0.88}\scriptsize
\caption{Results for every cell, and per victim (Mean: over the five datasets). A$\Delta$: attack-MAE change (\%).}
\label{tab:cells}
\setlength{\tabcolsep}{2.2pt}
\begin{tabular}{llcccccccccccccccc}
\toprule
 & & \multicolumn{8}{c}{\textsc{BackTime}~\cite{lin2024backtime}} & \multicolumn{8}{c}{\textsc{TDBA}~\cite{liu2026tdba}}  \\
\cmidrule(lr){3-10}\cmidrule(lr){11-18}
Dataset & Model & U & NT & ST & TG & A$\Delta$ST$\uparrow$ & A$\Delta$TG$\uparrow$ & $\Delta$ST$\downarrow$ & $\Delta$TG$\downarrow$ & U & NT & ST & TG & A$\Delta$ST$\uparrow$ & A$\Delta$TG$\uparrow$ & $\Delta$ST$\downarrow$ & $\Delta$TG$\downarrow$  \\
\midrule
\textsc{PEMS03} & FEDformer & 0.949 & 0.141 & \textbf{0.122} & 0.123 & $+264.4$ & $\mathbf{+309.7}$ & $\mathbf{-2.4}$ & $+4.1$ & 0.666 & 0.325 & \textbf{0.282} & 0.747 & $\mathbf{+46.6}$ & $+4.0$ & $\mathbf{-1.3}$ & $+4.5$ \\
 & Autoformer & 0.674 & 0.161 & \textbf{0.129} & 0.183 & $\mathbf{+96.4}$ & $+89.5$ & $\mathbf{-5.2}$ & $+1.0$ & 0.626 & 0.350 & \textbf{0.312} & 0.677 & $\mathbf{+34.0}$ & $+1.0$ & $\mathbf{-3.1}$ & $+3.8$ \\
 & TimesNet & 0.833 & 0.190 & 0.183 & \textbf{0.171} & $+95.1$ & $\mathbf{+99.0}$ & $\mathbf{-7.3}$ & $-3.2$ & 0.634 & 0.471 & \textbf{0.317} & 0.547 & $\mathbf{+20.4}$ & $+0.9$ & $\mathbf{-7.1}$ & $-2.5$ \\
 & Informer & 0.441 & 0.319 & \textbf{0.222} & 0.228 & $\mathbf{+32.1}$ & $+24.9$ & $\mathbf{-4.8}$ & $-2.7$ & 0.555 & 0.494 & \textbf{0.386} & 0.565 & $\mathbf{+17.9}$ & $-0.7$ & $\mathbf{-4.3}$ & $-0.5$ \\
 & PatchTST & 0.826 & 0.154 & \textbf{0.146} & 0.148 & $+103.0$ & $\mathbf{+104.3}$ & $\mathbf{-1.3}$ & $+8.1$ & 0.613 & 0.319 & \textbf{0.254} & 0.327 & $\mathbf{+41.4}$ & $+28.7$ & $\mathbf{-0.5}$ & $+9.9$ \\
 & iTransformer & 0.997 & 0.125 & \textbf{0.171} & 0.353 & $\mathbf{+1097.9}$ & $+704.5$ & $\mathbf{-0.4}$ & $+6.1$ & 0.591 & 0.277 & \textbf{0.351} & 0.710 & $\mathbf{+23.6}$ & $+10.1$ & $\mathbf{-0.8}$ & $+5.6$ \\
\midrule
\textsc{PEMS04} & FEDformer & 0.871 & 0.191 & 0.193 & \textbf{0.174} & $+155.5$ & $\mathbf{+192.0}$ & $\mathbf{-1.2}$ & $+3.8$ & 0.615 & 0.324 & \textbf{0.292} & 0.712 & $\mathbf{+35.3}$ & $-1.8$ & $\mathbf{-0.7}$ & $+3.9$ \\
 & Autoformer & 0.596 & 0.237 & \textbf{0.196} & 0.250 & $\mathbf{+64.3}$ & $+40.9$ & $\mathbf{+0.2}$ & $+4.1$ & 0.616 & 0.339 & \textbf{0.325} & 0.667 & $\mathbf{+27.2}$ & $-1.4$ & $\mathbf{-1.9}$ & $+5.6$ \\
 & TimesNet & 0.731 & 0.186 & 0.177 & \textbf{0.168} & $+77.5$ & $\mathbf{+77.8}$ & $\mathbf{-4.5}$ & $-3.4$ & 0.469 & 0.369 & \textbf{0.241} & 0.361 & $\mathbf{+20.2}$ & $+5.1$ & $\mathbf{-5.5}$ & $-2.6$ \\
 & Informer & 0.367 & 0.251 & \textbf{0.198} & 0.216 & $\mathbf{+30.0}$ & $+19.1$ & $\mathbf{-2.3}$ & $-0.1$ & 0.429 & 0.428 & \textbf{0.298} & 0.454 & $\mathbf{+16.2}$ & $-2.4$ & $\mathbf{-2.8}$ & $+1.0$ \\
 & PatchTST & 0.680 & 0.193 & \textbf{0.176} & 0.180 & $\mathbf{+74.2}$ & $+73.5$ & $\mathbf{+0.8}$ & $+8.3$ & 0.444 & 0.274 & \textbf{0.240} & 0.273 & $\mathbf{+26.3}$ & $+21.7$ & $\mathbf{+0.2}$ & $+9.1$ \\
 & iTransformer & 0.997 & 0.180 & \textbf{0.251} & 0.452 & $\mathbf{+965.1}$ & $+582.2$ & $\mathbf{-0.7}$ & $+5.3$ & 0.627 & 0.289 & \textbf{0.344} & 0.738 & $\mathbf{+24.5}$ & $+8.3$ & $\mathbf{-0.3}$ & $+5.8$ \\
\midrule
\textsc{PEMS08} & FEDformer & 0.925 & 0.151 & 0.167 & \textbf{0.138} & $+243.4$ & $\mathbf{+263.8}$ & $\mathbf{+1.1}$ & $+4.4$ & 0.600 & 0.330 & \textbf{0.278} & 0.745 & $\mathbf{+50.2}$ & $+0.8$ & $\mathbf{+1.4}$ & $+5.3$ \\
 & Autoformer & 0.614 & 0.175 & \textbf{0.152} & 0.176 & $\mathbf{+90.3}$ & $+73.6$ & $\mathbf{-0.3}$ & $+5.9$ & 0.626 & 0.373 & \textbf{0.339} & 0.707 & $\mathbf{+34.4}$ & $+0.5$ & $\mathbf{-0.4}$ & $+7.8$ \\
 & TimesNet & 0.774 & 0.231 & 0.216 & \textbf{0.163} & $+82.6$ & $\mathbf{+90.7}$ & $\mathbf{-2.9}$ & $-0.1$ & 0.491 & 0.419 & \textbf{0.289} & 0.433 & $\mathbf{+20.4}$ & $+3.4$ & $\mathbf{-4.4}$ & $+1.1$ \\
 & Informer & 0.349 & 0.305 & \textbf{0.225} & 0.249 & $\mathbf{+22.5}$ & $+10.4$ & $\mathbf{-4.6}$ & $-3.7$ & 0.435 & 0.452 & \textbf{0.322} & 0.447 & $\mathbf{+16.3}$ & $-5.2$ & $-1.9$ & $\mathbf{-2.2}$ \\
 & PatchTST & 0.821 & 0.299 & 0.212 & \textbf{0.175} & $\mathbf{+77.3}$ & $+74.2$ & $\mathbf{+3.9}$ & $+13.1$ & 0.503 & 0.309 & \textbf{0.268} & 0.354 & $\mathbf{+44.0}$ & $+23.9$ & $\mathbf{+3.3}$ & $+15.8$ \\
 & iTransformer & 0.995 & 0.140 & \textbf{0.202} & 0.561 & $\mathbf{+725.8}$ & $+359.1$ & $\mathbf{-0.3}$ & $+7.4$ & 0.620 & 0.279 & \textbf{0.330} & 0.765 & $\mathbf{+28.7}$ & $+10.0$ & $\mathbf{-1.8}$ & $+8.9$ \\
\midrule
\textsc{Weather} & FEDformer & 0.992 & 0.043 & \textbf{0.044} & 0.704 & $\mathbf{+704.4}$ & $+214.3$ & $\mathbf{-0.8}$ & $+21.0$ & 0.619 & 0.256 & \textbf{0.288} & 0.788 & $\mathbf{+64.6}$ & $+5.5$ & $\mathbf{-6.3}$ & $+37.9$ \\
 & Autoformer & 0.986 & 0.055 & \textbf{0.057} & 0.640 & $\mathbf{+299.5}$ & $+130.6$ & $\mathbf{-5.5}$ & $+10.7$ & 0.788 & 0.299 & \textbf{0.305} & 0.674 & $\mathbf{+50.3}$ & $+1.6$ & $\mathbf{-10.3}$ & $-3.6$ \\
 & TimesNet & 0.933 & 0.163 & \textbf{0.182} & 0.890 & $\mathbf{+205.7}$ & $+10.7$ & $\mathbf{-3.1}$ & $+6.3$ & 0.706 & 0.655 & \textbf{0.578} & 0.739 & $\mathbf{-19.3}$ & $-29.7$ & $\mathbf{-9.4}$ & $+4.0$ \\
 & Informer & 0.375 & 0.398 & \textbf{0.278} & 0.281 & $\mathbf{+19.5}$ & $+6.2$ & $-6.3$ & $\mathbf{-24.1}$ & 0.596 & 0.578 & \textbf{0.500} & 0.645 & $\mathbf{-4.0}$ & $-23.3$ & $-34.6$ & $\mathbf{-36.0}$ \\
 & PatchTST & 0.998 & 0.039 & \textbf{0.038} & 0.646 & $\mathbf{+1415.5}$ & $+813.6$ & $\mathbf{-2.3}$ & $+4.7$ & 0.603 & 0.171 & \textbf{0.172} & 0.708 & $\mathbf{+77.2}$ & $+5.1$ & $\mathbf{-10.6}$ & $+18.9$ \\
 & iTransformer & 0.993 & 0.039 & \textbf{0.070} & 0.993 & $\mathbf{+819.3}$ & $+37.5$ & $-10.2$ & $\mathbf{-13.4}$ & 0.559 & 0.415 & \textbf{0.272} & 0.576 & $+10.7$ & $\mathbf{+16.7}$ & $-5.7$ & $\mathbf{-8.7}$ \\
\midrule
\textsc{Electricity} & FEDformer & 0.981 & 0.047 & \textbf{0.013} & 0.031 & $\mathbf{+405.2}$ & $+354.5$ & $\mathbf{-24.3}$ & $+6.8$ & 0.314 & 0.266 & \textbf{0.297} & 0.335 & $+2.1$ & $\mathbf{+14.8}$ & $\mathbf{-1.8}$ & $+17.6$ \\
 & Autoformer & 0.993 & 0.029 & \textbf{0.035} & 0.527 & $\mathbf{+383.9}$ & $+150.0$ & $\mathbf{-27.1}$ & $-3.4$ & 0.350 & 0.258 & \textbf{0.297} & 0.368 & $+10.3$ & $\mathbf{+22.9}$ & $\mathbf{-5.9}$ & $+18.8$ \\
 & TimesNet & 0.902 & 0.014 & \textbf{0.004} & 0.084 & $\mathbf{+551.2}$ & $+346.5$ & $\mathbf{-46.9}$ & $+12.5$ & 0.285 & 0.251 & 0.285 & \textbf{0.274} & $+4.6$ & $\mathbf{+10.9}$ & $\mathbf{+2.8}$ & $+4.6$ \\
 & Informer & 0.350 & 0.049 & 0.053 & \textbf{0.034} & $\mathbf{+59.5}$ & $+42.9$ & $\mathbf{-4.6}$ & $+0.4$ & 0.530 & 0.476 & \textbf{0.477} & 0.521 & $\mathbf{+16.0}$ & $+5.3$ & $+2.2$ & $\mathbf{+1.2}$ \\
 & PatchTST & 0.920 & 0.018 & \textbf{0.005} & 0.026 & $\mathbf{+239.3}$ & $+134.9$ & $\mathbf{-42.9}$ & $+18.7$ & 0.327 & 0.304 & \textbf{0.317} & 0.318 & $+5.3$ & $\mathbf{+10.3}$ & $\mathbf{-0.7}$ & $+23.4$ \\
 & iTransformer & 0.979 & 0.010 & \textbf{0.025} & 0.368 & $\mathbf{+348.0}$ & $+110.7$ & $\mathbf{-35.4}$ & $+19.9$ & 0.372 & 0.311 & \textbf{0.340} & 0.365 & $+4.2$ & $\mathbf{+12.6}$ & $\mathbf{-5.4}$ & $+28.1$ \\
\midrule
Mean & FEDformer & 0.944 & 0.114 & \textbf{0.108} & 0.234 & $\mathbf{+354.6}$ & $+266.8$ & $\mathbf{-5.5}$ & $+8.0$ & 0.563 & 0.300 & \textbf{0.287} & 0.666 & $\mathbf{+39.8}$ & $+4.7$ & $\mathbf{-1.8}$ & $+13.9$ \\
 & Autoformer & 0.773 & 0.131 & \textbf{0.114} & 0.355 & $\mathbf{+186.9}$ & $+96.9$ & $\mathbf{-7.6}$ & $+3.6$ & 0.601 & 0.324 & \textbf{0.316} & 0.618 & $\mathbf{+31.2}$ & $+4.9$ & $\mathbf{-4.3}$ & $+6.5$ \\
 & TimesNet & 0.835 & 0.157 & \textbf{0.152} & 0.295 & $\mathbf{+202.4}$ & $+124.9$ & $\mathbf{-12.9}$ & $+2.4$ & 0.517 & 0.433 & \textbf{0.342} & 0.471 & $\mathbf{+9.2}$ & $-1.9$ & $\mathbf{-4.7}$ & $+0.9$ \\
 & Informer & 0.376 & 0.264 & \textbf{0.196} & 0.202 & $\mathbf{+32.7}$ & $+20.7$ & $-4.5$ & $\mathbf{-6.0}$ & 0.509 & 0.485 & \textbf{0.397} & 0.526 & $\mathbf{+12.5}$ & $-5.3$ & $\mathbf{-8.3}$ & $-7.3$ \\
 & PatchTST & 0.849 & 0.140 & \textbf{0.115} & 0.235 & $\mathbf{+381.8}$ & $+240.1$ & $\mathbf{-8.4}$ & $+10.6$ & 0.498 & 0.275 & \textbf{0.250} & 0.396 & $\mathbf{+38.8}$ & $+18.0$ & $\mathbf{-1.7}$ & $+15.4$ \\
 & iTransformer & 0.992 & 0.099 & \textbf{0.144} & 0.545 & $\mathbf{+791.2}$ & $+358.8$ & $\mathbf{-9.4}$ & $+5.1$ & 0.554 & 0.314 & \textbf{0.327} & 0.631 & $\mathbf{+18.3}$ & $+11.5$ & $\mathbf{-2.8}$ & $+7.9$ \\
\bottomrule
\end{tabular}}
\end{table*}
\section{Results per Architecture and per Cell}
\label{app:arch}

Throughout the appendix, U or Undef.\ denotes undefended; No-trig.\ or NT, the undefended no-trigger rate; ST,
\textsc{SafeTime}; TG, \textsc{TimeGuard}; $I$ or Impl., the implanted effect; Found and False, the attacked and
unattacked channels flagged; AST and ATG, the attack MAE with \textsc{SafeTime} and \textsc{TimeGuard}; $\Delta$, the
clean-MAE change (\%); Ratio, the clean MAE over the undefended model's. Bold marks the better of ST and TG.

Table~\ref{tab:cells} reports Table~\ref{tab:main} per cell and per victim. Under \textsc{BackTime}, \textsc{SafeTime}
has a lower ASR than \textsc{TimeGuard} on every architecture. Informer barely responds to the trigger even after
poisoning, and iTransformer responds to the \textsc{BackTime} trigger even when trained on the clean series, which no
relabelling can remove. Table~\ref{tab:victims} retrains the \textsc{DEBA-MTS}, \textsc{IFBA}, adaptive, and direction
poisons on the other architectures: undefended, they implant up to $0.998$, and on \textsc{SafeTime}'s labels every
model keeps at most $0.057$ except \textsc{IFBA} on iTransformer ($0.021$ to $0.124$). The clean error falls in $46$ of
the $72$ \textsc{DEBA-MTS}, \textsc{IFBA}, and adaptive models and rises by more than $5\%$ in $17$, mostly on
\textsc{Weather} under \textsc{DEBA-MTS} and on \textsc{PEMS08} and \textsc{Electricity} under \textsc{IFBA}.
Table~\ref{tab:signflip} reports every sign-changing poison, and Table~\ref{tab:redundancy} the cross-channel redundancy
on which relabelling relies: $4$ of the $21$ \textsc{Weather} channels and at most $2\%$ of the channels elsewhere fall
below $R^2 = 0.5$.

\section{Stronger-Attacker Results}
\label{app:attackers}

In Table~\ref{tab:stealth}, all $15$ stealthier \textsc{BackTime} settings implant $0.318$ to $0.984$; \textsc{SafeTime}
lowers the ASR to at most $0.172$, whereas \textsc{TimeGuard} leaves an ASR above $0.4$ in four. A single poisoned
channel is flagged in all $15$ poisons (Table~\ref{tab:onechannel}), although without the within-channel test of
Table~\ref{tab:ablation} only $1$ of the $14$ in the ablation is flagged. With two or three simultaneous attackers
(Table~\ref{tab:multi}), every attacked channel is found and each attacker is left an ASR of at most $0.189$.

Table~\ref{tab:kn-rest} lists the remaining knowledgeable-attacker combinations that implant; on \textsc{PEMS04}, the C R combination
is flagged on only $11$ of $31$ channels. On the other five victims (Table~\ref{tab:transfer}), the \textsc{Weather}
poison stays at an ASR of at most $0.04$ under \textsc{SafeTime}, against up to $0.995$ under \textsc{TimeGuard}; on
iTransformer, whose clean-trained models already respond to the triggers, \textsc{SafeTime} leaves $0.51$ on the
\textsc{PEMS03} combination. Under realistic noise (Table~\ref{tab:noise}), \textsc{SafeTime} flags all $107$ attacked
\textsc{PEMS03} channels at every level and placement; at $\rho = 1$, its mean ASR is $0.144$ and $0.334$, against
$0.440$ and $0.572$ for \textsc{TimeGuard}, whose \textsc{PEMS03} ASR under \textsc{BackTime} rises from $0.139$ to
$0.873$ as the noise grows.

When \textsc{BackTime} attacks half of the channels, the edge of the minority that the threat model assumes
(Section~\ref{sec:threat-model}), \textsc{SafeTime} finds at least $98.8\%$ of them and leaves at most $0.093$, except
PatchTST on \textsc{Electricity} ($1$ of $160$ found), and less than \textsc{TimeGuard} on average on every dataset
(Table~\ref{tab:fraction}). \textsc{TimeGuard} leaves $0.432$ to $0.969$ on \textsc{Weather} for every victim except Informer, $0.468$ to
$0.842$ on \textsc{Electricity} except Informer and PatchTST, and up to $0.475$ on the traffic datasets.

\setcounter{table}{17}
\begin{table*}[!t]\centering{\renewcommand{\arraystretch}{0.88}
\caption{Victim architectures: \textsc{DEBA-MTS}, \textsc{IFBA}, the adaptive poison (outcome goal, feedback), and the direction poisons ($f$: with feedback), retrained on every victim. ASR: with \textsc{SafeTime}.}
\label{tab:victims}
\scriptsize
\setlength{\tabcolsep}{1.8pt}
\begin{tabular}{@{}llrrrrrrrrrrrrrrrrrr@{}}
\toprule
 & & \multicolumn{3}{c}{FEDformer} & \multicolumn{3}{c}{Autoformer} & \multicolumn{3}{c}{TimesNet} & \multicolumn{3}{c}{Informer} & \multicolumn{3}{c}{PatchTST} & \multicolumn{3}{c}{iTransformer} \\
\cmidrule(lr){3-5}\cmidrule(lr){6-8}\cmidrule(lr){9-11}\cmidrule(lr){12-14}\cmidrule(lr){15-17}\cmidrule(lr){18-20}
Attack & Dataset & $I$ U & $I$ ST$\downarrow$ & ASR$\downarrow$ & $I$ U & $I$ ST$\downarrow$ & ASR$\downarrow$ & $I$ U & $I$ ST$\downarrow$ & ASR$\downarrow$ & $I$ U & $I$ ST$\downarrow$ & ASR$\downarrow$ & $I$ U & $I$ ST$\downarrow$ & ASR$\downarrow$ & $I$ U & $I$ ST$\downarrow$ & ASR$\downarrow$  \\
\midrule
\textsc{DEBA-MTS} & \textsc{PEMS03} & $.795$ & $-.001$ & $.144$ & $.440$ & $.002$ & $.134$ & $.501$ & $.002$ & $.199$ & $.130$ & $.000$ & $.213$ & $.579$ & $.001$ & $.153$ & $.860$ & $.038$ & $.167$ \\
 & \textsc{PEMS04} & $.600$ & $.030$ & $.255$ & $.344$ & $.022$ & $.245$ & $.141$ & $.000$ & $.202$ & $-.012$ & $.000$ & $.218$ & $-.022$ & $.001$ & $.213$ & $.784$ & $.055$ & $.259$ \\
 & \textsc{PEMS08} & $.779$ & $.001$ & $.150$ & $.422$ & $.004$ & $.139$ & $.293$ & $.001$ & $.209$ & $-.027$ & $.002$ & $.225$ & $.049$ & $.001$ & $.183$ & $.853$ & $.054$ & $.182$ \\
 & \textsc{Weather} & $.989$ & $.002$ & $.015$ & $.986$ & $-.010$ & $.018$ & $.797$ & $.000$ & $.015$ & $.280$ & $.000$ & $.015$ & $.989$ & $.000$ & $.026$ & $.988$ & $.002$ & $.015$ \\
 & \textsc{Electricity} & $.998$ & $.001$ & $.003$ & $.932$ & $.001$ & $.003$ & $.989$ & $.000$ & $.002$ & $.851$ & $.000$ & $.003$ & $.988$ & $.000$ & $.001$ & $.996$ & $.001$ & $.005$ \\
\midrule
\textsc{IFBA} & \textsc{PEMS03} & $.639$ & $.034$ & $.281$ & $.409$ & $.044$ & $.276$ & $.314$ & $.003$ & $.310$ & $.026$ & $.003$ & $.360$ & $.462$ & $-.000$ & $.258$ & $.526$ & $.078$ & $.293$ \\
 & \textsc{PEMS04} & $.577$ & $.051$ & $.309$ & $.357$ & $.057$ & $.338$ & $.080$ & $.002$ & $.280$ & $-.005$ & $.001$ & $.302$ & $.253$ & $.004$ & $.292$ & $.477$ & $.124$ & $.345$ \\
 & \textsc{PEMS08} & $.654$ & $.031$ & $.338$ & $.457$ & $.025$ & $.357$ & $.171$ & $.001$ & $.308$ & $-.008$ & $-.001$ & $.302$ & $.413$ & $.002$ & $.261$ & $.516$ & $.122$ & $.348$ \\
 & \textsc{Weather} & $.961$ & $-.001$ & $.039$ & $.946$ & $.000$ & $.050$ & $.833$ & $-.008$ & $.166$ & $.037$ & $-.005$ & $.255$ & $.962$ & $.001$ & $.045$ & $.958$ & $.021$ & $.055$ \\
 & \textsc{Electricity} & $.911$ & $.028$ & $.098$ & $.478$ & $.039$ & $.101$ & $.854$ & $-.007$ & $.118$ & $.268$ & $.008$ & $.104$ & $.883$ & $.003$ & $.098$ & $.888$ & $.083$ & $.123$ \\
\midrule
Adaptive & \textsc{PEMS03} & $.725$ & $.003$ & $.155$ & $.412$ & $-.008$ & $.132$ & $.472$ & $.001$ & $.215$ & $.007$ & $-.001$ & $.256$ & $.554$ & $.000$ & $.164$ & $.836$ & $.039$ & $.188$ \\
 & \textsc{Electricity} & $.938$ & $.009$ & $.019$ & $.892$ & $.011$ & $.022$ & $.914$ & $.010$ & $.024$ & $.207$ & $.006$ & $.040$ & $.933$ & $.006$ & $.019$ & $.967$ & $.034$ & $.046$ \\
\midrule
Direction & \textsc{PEMS03} & $.894$ & $.017$ & $.057$ & $.390$ & $.018$ & $.069$ & $.671$ & $.001$ & $.077$ & $-.003$ & $.000$ & $.079$ & $-.016$ & $.001$ & $.066$ & $.939$ & $.042$ & $.083$ \\
 & \textsc{Electricity} & $.937$ & $.007$ & $.009$ & $.804$ & $.008$ & $.010$ & $.978$ & $.008$ & $.011$ & $.032$ & $.007$ & $.014$ & $.878$ & $.006$ & $.007$ & $.921$ & $.016$ & $.019$ \\
Direction$^{f}$ & \textsc{PEMS03} & $.870$ & $.018$ & $.057$ & $.438$ & $.016$ & $.068$ & $.644$ & $.001$ & $.077$ & $-.004$ & $.000$ & $.077$ & $-.012$ & $.001$ & $.066$ & $.945$ & $.041$ & $.084$ \\
 & \textsc{Electricity} & $.932$ & $.007$ & $.009$ & $.814$ & $.009$ & $.011$ & $.972$ & $.009$ & $.011$ & $.030$ & $.006$ & $.013$ & $.856$ & $.006$ & $.008$ & $.919$ & $.016$ & $.019$ \\
\bottomrule
\end{tabular}}
\end{table*}
\setcounter{table}{18}
\begin{table*}[!t]\centering{\renewcommand{\arraystretch}{0.88}
\caption{Victim training loss: FEDformer victims trained with Smooth L1 (the loss of every other result), squared error, and L1 ($\pm$ standard deviation over training seeds).}
\label{tab:losses}
\scriptsize
\setlength{\tabcolsep}{2.4pt}
\begin{tabular}{@{}llccccccccc@{}}
\toprule
 & & \multicolumn{3}{c}{Smooth L1} & \multicolumn{3}{c}{Squared error} & \multicolumn{3}{c}{L1} \\
\cmidrule(lr){3-5}\cmidrule(lr){6-8}\cmidrule(lr){9-11}
Attack & Dataset & $I$ U & $I$ ST$\downarrow$ & $\Delta\downarrow$ & $I$ U & $I$ ST$\downarrow$ & $\Delta\downarrow$ & $I$ U & $I$ ST$\downarrow$ & $\Delta\downarrow$ \\
\midrule
\textsc{BackTime}~\cite{lin2024backtime} & \textsc{PEMS03} & $0.808_{\pm 0.004}$ & $0.010_{\pm 0.000}$ & $-2.4_{\pm 0.3}$ & $0.783_{\pm 0.008}$ & $0.003_{\pm 0.002}$ & $-2.9_{\pm 0.6}$ & $0.817_{\pm 0.010}$ & $0.008_{\pm 0.001}$ & $-1.9_{\pm 1.2}$ \\
 & \textsc{PEMS04} & $0.680_{\pm 0.000}$ & $0.014_{\pm 0.003}$ & $-1.2_{\pm 0.6}$ & $0.646_{\pm 0.011}$ & $0.010_{\pm 0.001}$ & $-1.0_{\pm 0.2}$ & $0.683_{\pm 0.002}$ & $0.015_{\pm 0.001}$ & $-1.2_{\pm 0.3}$ \\
 & \textsc{PEMS08} & $0.774_{\pm 0.011}$ & $0.004_{\pm 0.002}$ & $+1.1_{\pm 0.2}$ & $0.732_{\pm 0.010}$ & $-0.001_{\pm 0.002}$ & $+0.9_{\pm 0.7}$ & $0.776_{\pm 0.003}$ & $0.002_{\pm 0.001}$ & $+0.6_{\pm 0.2}$ \\
 & \textsc{Weather} & $0.949_{\pm 0.002}$ & $0.004_{\pm 0.001}$ & $-0.8_{\pm 6.1}$ & $0.863_{\pm 0.042}$ & $0.018_{\pm 0.042}$ & $-19.8_{\pm 5.3}$ & $0.944_{\pm 0.005}$ & $0.002_{\pm 0.001}$ & $-10.7_{\pm 3.8}$ \\
 & \textsc{Electricity} & $0.934_{\pm 0.015}$ & $0.009_{\pm 0.000}$ & $-24.3_{\pm 0.7}$ & $0.951_{\pm 0.018}$ & $0.011_{\pm 0.002}$ & $-28.9_{\pm 4.7}$ & $0.933_{\pm 0.013}$ & $0.009_{\pm 0.000}$ & $-23.1_{\pm 2.0}$ \\
\midrule
\textsc{TDBA}~\cite{liu2026tdba} & \textsc{PEMS03} & $0.341_{\pm 0.021}$ & $0.032_{\pm 0.017}$ & $-1.3_{\pm 0.3}$ & $0.248_{\pm 0.008}$ & $0.043_{\pm 0.011}$ & $-0.9_{\pm 0.8}$ & $0.339_{\pm 0.009}$ & $0.032_{\pm 0.013}$ & $-1.8_{\pm 0.2}$ \\
 & \textsc{PEMS04} & $0.292_{\pm 0.005}$ & $0.020_{\pm 0.004}$ & $-0.7_{\pm 0.6}$ & $0.261_{\pm 0.011}$ & $0.024_{\pm 0.010}$ & $-0.7_{\pm 0.6}$ & $0.306_{\pm 0.020}$ & $0.021_{\pm 0.001}$ & $-0.8_{\pm 0.3}$ \\
 & \textsc{PEMS08} & $0.270_{\pm 0.041}$ & $0.018_{\pm 0.020}$ & $+1.4_{\pm 1.3}$ & $0.237_{\pm 0.025}$ & $0.044_{\pm 0.012}$ & $+1.3_{\pm 0.5}$ & $0.250_{\pm 0.056}$ & $0.015_{\pm 0.030}$ & $+1.4_{\pm 0.2}$ \\
 & \textsc{Weather} & $0.362_{\pm 0.141}$ & $0.078_{\pm 0.066}$ & $-6.3_{\pm 0.9}$ & $0.321_{\pm 0.075}$ & $0.230_{\pm 0.207}$ & $-6.0_{\pm 10.4}$ & $0.436_{\pm 0.041}$ & $0.035_{\pm 0.065}$ & $-3.3_{\pm 15.7}$ \\
 & \textsc{Electricity} & $0.048_{\pm 0.002}$ & $0.034_{\pm 0.004}$ & $-1.8_{\pm 0.3}$ & $0.045_{\pm 0.006}$ & $0.020_{\pm 0.009}$ & $-9.8_{\pm 3.4}$ & $0.044_{\pm 0.001}$ & $0.038_{\pm 0.005}$ & $-1.9_{\pm 2.0}$ \\
\bottomrule
\end{tabular}}
\end{table*}
\setcounter{table}{19}
\begin{table*}[!t]\centering{\renewcommand{\arraystretch}{0.88}\scriptsize
\caption{Implanted effect left by \textsc{TimeGuard} (TG) and \textsc{SafeTime} (ST) when \textsc{BackTime} attacks $50\%$ of the channels. Bold: the lower of the two, negative effects counted as zero. $^\circ$: \textsc{SafeTime} finds fewer than half of the attacked channels.}
\label{tab:fraction}
\setlength{\tabcolsep}{4pt}
\begin{tabular}{@{}lrrrrrrrrrr@{}}
\toprule
 & \multicolumn{2}{c}{\textsc{PEMS03}} & \multicolumn{2}{c}{\textsc{PEMS04}} & \multicolumn{2}{c}{\textsc{PEMS08}} & \multicolumn{2}{c}{\textsc{Weather}} & \multicolumn{2}{c}{\textsc{Electricity}} \\
\cmidrule(lr){2-3}\cmidrule(lr){4-5}\cmidrule(lr){6-7}\cmidrule(lr){8-9}\cmidrule(lr){10-11}
Victim & TG & ST & TG & ST & TG & ST & TG & ST & TG & ST \\
\midrule
FEDformer & .034 & \textbf{.015} & .020 & \textbf{.013} & .016 & \textbf{.009} & .432 & \textbf{$-$.003} & .611 & \textbf{.013} \\
Autoformer & .013 & \textbf{.001} & .029 & \textbf{$-$.005} & .092 & \textbf{.019} & .923 & \textbf{.006} & .468 & \textbf{.007} \\
TimesNet & \textbf{$-$.004} & .002 & \textbf{$-$.004} & .004 & \textbf{$-$.012} & .004 & .807 & \textbf{.000} & .520 & \textbf{.000} \\
Informer & .001 & .001 & $-$.009 & $-$.001 & \textbf{$-$.002} & .001 & .032 & \textbf{.000} & .005 & \textbf{$-$.001} \\
PatchTST & $-$.001 & .000 & .002 & .002 & .003 & .003 & .466 & \textbf{.000} & \textbf{.007} & .949$^\circ$ \\
iTransformer & .059 & \textbf{.036} & .236 & \textbf{.045} & .475 & \textbf{.093} & .969 & \textbf{.000} & .842 & \textbf{.015} \\
\bottomrule
\end{tabular}}
\end{table*}
\setcounter{table}{20}
\begin{table*}[!t]\centering{\renewcommand{\arraystretch}{0.88}\scriptsize
\caption{Knowledgeable-attacker and single-channel poisons (mean over attacker poisons) on the other victims. Each entry: U / ST / TG ASR (first line) and clean-MAE change of ST / TG (second).}
\label{tab:transfer}
\setlength{\tabcolsep}{4pt}
\begin{tabular}{lllllll}
\toprule
Dataset & Poison & Autoformer & TimesNet & Informer & PatchTST & iTransformer \\
\midrule
\textsc{PEMS03} & C R S & 0.54/\textbf{0.12}/0.20 & 0.24/\textbf{0.17}/\textbf{0.17} & 0.25/0.22/\textbf{0.21} & 0.21/\textbf{0.13}/0.14 & 0.67/\textbf{0.51}/0.82 \\
 & clean $\Delta$MAE & \textbf{-0.2}/+14.9 & \textbf{-0.9}/+0.3 & \textbf{+0.3}/+1.0 & \textbf{-0.1}/+10.6 & \textbf{-0.4}/+6.9 \\
\textsc{Weather} & C R S & 0.94/\textbf{0.03}/0.45 & 0.57/\textbf{0.02}/0.11 & 0.03/\textbf{0.02}/\textbf{0.02} & 0.05/\textbf{0.02}/0.03 & 0.99/\textbf{0.04}/0.99 \\
 & clean $\Delta$MAE & -9.2/\textbf{-14.8} & \textbf{-2.2}/+7.1 & -2.1/\textbf{-38.2} & \textbf{+4.0}/+10.6 & +10.7/\textbf{+4.5} \\
\textsc{PEMS03} & one channel & 0.58/\textbf{0.10}/0.12 & 0.12/0.09/\textbf{0.05} & 0.24/0.14/\textbf{0.10} & 0.10/0.07/\textbf{0.05} & 0.68/\textbf{0.54}/0.82 \\
 & clean $\Delta$MAE & \textbf{+1.9}/+13.5 & +1.3/\textbf{-0.6} & +1.7/\textbf{+1.0} & \textbf{+1.3}/+10.7 & \textbf{+0.0}/+7.7 \\
\textsc{PEMS08} & one channel & 0.49/\textbf{0.16}/0.44 & 0.33/0.25/\textbf{0.20} & 0.28/\textbf{0.19}/0.22 & 0.17/0.17/\textbf{0.15} & 0.70/\textbf{0.64}/0.85 \\
 & clean $\Delta$MAE & \textbf{-2.1}/+10.1 & \textbf{-0.1}/+6.6 & \textbf{-0.6}/+1.9 & \textbf{-0.1}/+11.9 & \textbf{-0.3}/+8.4 \\
\textsc{Weather} & one channel & 0.30/\textbf{0.06}/0.08 & 0.56/\textbf{0.05}/\textbf{0.05} & 0.09/\textbf{0.05}/\textbf{0.05} & 0.09/\textbf{0.05}/\textbf{0.05} & 0.99/\textbf{0.06}/0.97 \\
 & clean $\Delta$MAE & +4.4/\textbf{-22.5} & \textbf{-6.4}/+15.4 & +18.1/\textbf{-10.7} & \textbf{+1.5}/+8.2 & +12.7/\textbf{+5.8} \\
\textsc{Electricity} & one channel & 0.63/\textbf{0.00}/0.61 & 0.14/\textbf{0.00}/\textbf{0.00} & 0.14/\textbf{0.00}/\textbf{0.00} & 0.12/\textbf{0.00}/\textbf{0.00} & 0.50/\textbf{0.38}/0.76 \\
 & clean $\Delta$MAE & \textbf{+0.5}/+30.9 & \textbf{-0.4}/+23.9 & -0.3/\textbf{-1.4} & \textbf{-0.4}/+54.3 & \textbf{-1.4}/+38.6 \\
\bottomrule
\end{tabular}}
\end{table*}
\setcounter{table}{21}
\begin{table*}[!t]\centering{\renewcommand{\arraystretch}{0.88}
\caption{Stronger fixed attackers: stealthier \textsc{BackTime} settings, one poisoned channel, several simultaneous attackers, realistic noise on every planted copy, and the further knowledgeable-attacker combinations that implant or that Section~\ref{sec:adaptive} discusses (the other $11$, and the three on \textsc{PEMS08}, do not implant) (each trigger and target receives its own block of natural short-scale variation, scaled by $\rho$; per dataset: mean over the six victims, Found and False summed over them). Ratio: clean MAE of \textsc{SafeTime} over undefended (several attackers: of the whole model).}
\label{tab:stealth}
\label{tab:onechannel}
\label{tab:multi}
\label{tab:kn-rest}
\label{tab:noise}
\scriptsize
\setlength{\tabcolsep}{5pt}
\begin{tabular}{llrrcccccccc}
\toprule
Dataset & Setting & Found$\uparrow$ & False$\downarrow$ & Undef. & No-trig. & ST & TG & \multicolumn{3}{c}{Attack MAE$\uparrow$ (U / ST / TG)} & Ratio$\downarrow$  \\
\midrule
\multicolumn{12}{@{}l}{\emph{Stealthier settings, one changed at a time}} \\
\textsc{PEMS03} & $10\%$ channels & 36/36 & 2 & 0.739 & 0.161 & \textbf{0.129} & 0.152 & 18.3 & \textbf{37.9} & 35.5 & 0.99 \\
 & $5\%$ channels & 18/18 & 19 & 0.809 & 0.202 & \textbf{0.132} & 0.174 & 17.2 & \textbf{39.3} & 33.4 & 1.02 \\
 & $1\%$ rate & 107/107 & 1 & 0.809 & 0.128 & \textbf{0.126} & 0.170 & 17.2 & \textbf{38.6} & 31.3 & 1.00 \\
 & magnitude penalty $\times4$ & 107/107 & 1 & 0.928 & 0.135 & \textbf{0.130} & 0.157 & 11.6 & \textbf{37.5} & 34.5 & 0.98 \\
 & trigger length 2 & 107/107 & 1 & 0.895 & 0.140 & 0.122 & \textbf{0.107} & 12.8 & 36.3 & \textbf{39.5} & 0.99 \\
\textsc{Weather} & $10\%$ channels & 2/2 & 1 & 0.966 & 0.043 & \textbf{0.033} & 0.496 & 25.0 & \textbf{90.2} & 49.1 & 1.01 \\
 & $5\%$ channels & 1/1 & 3 & 0.381 & 0.062 & \textbf{0.064} & 0.089 & 49.4 & \textbf{93.9} & 61.2 & 1.02 \\
 & $1\%$ rate & 6/6 & 3 & 0.511 & 0.043 & \textbf{0.172} & 0.332 & 47.1 & \textbf{98.8} & 58.9 & 1.83 \\
 & magnitude penalty $\times4$ & 6/6 & 1 & 0.992 & 0.043 & \textbf{0.047} & 0.642 & 8.2 & \textbf{92.3} & 40.9 & 0.94 \\
 & trigger length 2 & 6/6 & 2 & 0.982 & 0.048 & \textbf{0.049} & 0.478 & 13.0 & \textbf{90.5} & 50.6 & 1.00 \\
\textsc{Electricity} & $10\%$ channels & 31/32 & 3 & 0.977 & 0.014 & \textbf{0.034} & 0.044 & 1619 & \textbf{4874} & 4801 & 0.93 \\
 & $5\%$ channels & 16/16 & 2 & 0.945 & 0.015 & \textbf{0.003} & \textbf{0.003} & 753.5 & \textbf{3578} & 3210 & 0.96 \\
 & $1\%$ rate & 91/96 & 3 & 0.925 & 0.002 & 0.042 & \textbf{0.017} & 1237 & \textbf{4074} & 3603 & 0.98 \\
 & magnitude penalty $\times4$ & 95/96 & 2 & 0.988 & 0.016 & \textbf{0.012} & 0.794 & 1112 & \textbf{4176} & 1861 & 0.77 \\
 & trigger length 2 & 95/96 & 18 & 0.989 & 0.006 & \textbf{0.011} & 0.018 & 818.5 & 4212 & \textbf{4818} & 0.86 \\
\midrule
\multicolumn{12}{@{}l}{\emph{One poisoned channel, mean over three attacker poisons}} \\
\textsc{PEMS03} & one channel & 3/3 & 8.7 & 0.339 & 0.100 & 0.151 & \textbf{0.119} & 24.1 & \textbf{30.7} & 29.4 & \textbf{1.01} \\
\textsc{PEMS04} & one channel & 3/3 & 0.0 & 0.217 & 0.087 & 0.143 & \textbf{0.126} & 29.0 & \textbf{32.7} & 31.0 & \textbf{1.00} \\
\textsc{PEMS08} & one channel & 3/3 & 3.7 & 0.405 & 0.124 & \textbf{0.180} & 0.189 & 26.0 & \textbf{35.5} & 32.5 & \textbf{1.00} \\
\textsc{Weather} & one channel & 3/3 & 3.7 & 0.147 & 0.021 & \textbf{0.021} & 0.030 & 52.6 & \textbf{86.2} & 66.9 & \textbf{0.93} \\
\textsc{Electricity} & one channel & 3/3 & 3.3 & 0.538 & 0.029 & 0.134 & \textbf{0.001} & 1846 & 2550 & \textbf{2799} & \textbf{1.01} \\
\midrule
\multicolumn{12}{@{}l}{\emph{Several attackers on disjoint channel groups (attacker $j$ of $n$)}} \\
\textsc{PEMS03} & attacker 1 of 2 & 108/108 & -- & 0.910 & 0.158 & 0.170 & \textbf{0.133} & 12.9 & 39.2 & \textbf{51.7} & 1.02 \\
 & attacker 2 of 2 &  &  & 0.926 & 0.121 & \textbf{0.167} & 0.191 & 12.0 & \textbf{30.2} & 28.9 &  \\
 & attacker 1 of 3 & 108/108 & -- & 0.832 & 0.195 & 0.174 & \textbf{0.142} & 16.0 & 39.1 & \textbf{41.2} & 1.02 \\
 & attacker 2 of 3 &  &  & 0.830 & 0.181 & 0.165 & \textbf{0.164} & 18.0 & \textbf{41.9} & 39.5 &  \\
 & attacker 3 of 3 &  &  & 0.722 & 0.142 & \textbf{0.189} & 0.204 & 17.6 & \textbf{31.3} & 29.8 &  \\
\textsc{Electricity} & attacker 1 of 2 & 96/96 & -- & 0.986 & 0.045 & \textbf{0.001} & 0.003 & 426.4 & \textbf{3519} & 2997 & 0.77 \\
 & attacker 2 of 2 &  &  & 0.963 & 0.005 & \textbf{0.002} & 0.020 & 710.7 & \textbf{2826} & 2746 &  \\
 & attacker 1 of 3 & 96/96 & -- & 1.000 & 0.050 & \textbf{0.000} & \textbf{0.000} & 360.4 & \textbf{3484} & 2832 & 0.76 \\
 & attacker 2 of 3 &  &  & 0.972 & 0.062 & \textbf{0.010} & 0.030 & 635.1 & \textbf{4696} & 3688 &  \\
 & attacker 3 of 3 &  &  & 0.952 & 0.001 & \textbf{0.002} & 0.023 & 585.4 & 2608 & \textbf{2720} &  \\
\textsc{Weather} & attacker 1 of 2 & 6/6 & -- & 1.000 & 0.000 & \textbf{0.000} & \textbf{0.000} & 4.3 & \textbf{80.8} & 74.3 & 1.04 \\
 & attacker 2 of 2 &  &  & 0.029 & 0.024 & \textbf{0.025} & 0.027 & 49.5 & \textbf{58.7} & 57.5 &  \\
 & attacker 1 of 3 & 6/6 & -- & 1.000 & 0.000 & \textbf{0.000} & \textbf{0.000} & 16.3 & \textbf{85.9} & 83.7 & 0.88 \\
 & attacker 2 of 3 &  &  & 0.959 & 0.048 & \textbf{0.030} & 0.495 & 23.4 & \textbf{110.4} & 60.7 &  \\
 & attacker 3 of 3 &  &  & 0.000 & 0.000 & \textbf{0.000} & \textbf{0.000} & 45.5 & 55.5 & \textbf{57.6} &  \\
\midrule
\multicolumn{12}{@{}l}{\emph{Realistic noise on every copy, \textsc{BackTime} ($\rho = 1$, six victims)}} \\
\textsc{PEMS03} & noise $\rho = 1$ & 642/642 & 6 & 0.768 & 0.178 & \textbf{0.161} & 0.411 & 18.7 & \textbf{39.4} & 29.2 & 0.97 \\
\textsc{PEMS04} & noise $\rho = 1$ & 552/552 & 0 & 0.695 & 0.204 & \textbf{0.198} & 0.435 & 26.1 & \textbf{47.7} & 35.2 & 0.99 \\
\textsc{PEMS08} & noise $\rho = 1$ & 306/306 & 6 & 0.722 & 0.214 & \textbf{0.196} & 0.359 & 22.8 & \textbf{43.1} & 33.8 & 0.99 \\
\textsc{Weather} & noise $\rho = 1$ & 36/36 & 12 & 0.619 & 0.125 & \textbf{0.143} & 0.537 & 61.0 & \textbf{107.5} & 64.9 & 1.00 \\
\textsc{Electricity} & noise $\rho = 1$ & 572/576 & 36 & 0.486 & 0.014 & \textbf{0.022} & 0.459 & 3207 & \textbf{5108} & 3429 & 0.82 \\
All &  & 2{,}108/2{,}112 & 60 & 0.658 & 0.147 & \textbf{0.144} & 0.440 &  &  &  & 0.96 \\
\multicolumn{12}{@{}l}{\quad\emph{\textsc{PEMS03}, FEDformer: noise level and placement}} \\
\textsc{PEMS03} & noise $\rho = 0.5$ & 107/107 & 1 & 0.942 & 0.139 & \textbf{0.115} & 0.139 & 11.1 & \textbf{38.8} & 38.1 & 0.97 \\
 & noise $\rho = 1$ & 107/107 & 1 & 0.940 & 0.138 & \textbf{0.116} & 0.487 & 11.1 & \textbf{38.9} & 22.5 & 0.98 \\
 & noise $\rho = 2$ & 107/107 & 1 & 0.923 & 0.129 & \textbf{0.110} & 0.873 & 12.1 & \textbf{39.8} & 15.2 & 0.98 \\
 & noise on target only & 107/107 & 1 & 0.942 & 0.133 & \textbf{0.115} & 0.151 & 11.0 & \textbf{39.0} & 34.9 & 0.98 \\
 & noise on trigger only & 107/107 & 1 & 0.946 & 0.140 & \textbf{0.120} & 0.131 & 10.7 & 38.6 & \textbf{41.3} & 0.97 \\
\midrule
\multicolumn{12}{@{}l}{\emph{Realistic noise on every copy, \textsc{TDBA} ($\rho = 1$, six victims)}} \\
\textsc{PEMS03} & noise $\rho = 1$ & 642/642 & 6 & 0.604 & 0.365 & \textbf{0.324} & 0.629 & 21.6 & \textbf{27.5} & 22.3 & 0.97 \\
\textsc{PEMS04} & noise $\rho = 1$ & 552/552 & 0 & 0.523 & 0.329 & \textbf{0.303} & 0.561 & 30.6 & \textbf{37.1} & 30.8 & 0.98 \\
\textsc{PEMS08} & noise $\rho = 1$ & 306/306 & 6 & 0.545 & 0.355 & \textbf{0.319} & 0.606 & 26.2 & \textbf{33.0} & 26.4 & 0.99 \\
\textsc{Weather} & noise $\rho = 1$ & 36/36 & 12 & 0.662 & 0.365 & \textbf{0.399} & 0.688 & 69.4 & \textbf{60.0} & 54.7 & 0.91 \\
\textsc{Electricity} & noise $\rho = 1$ & 555/576 & 27 & 0.360 & 0.306 & \textbf{0.325} & 0.373 & 1713 & 1813 & \textbf{1836} & 0.97 \\
All &  & 2{,}091/2{,}112 & 51 & 0.539 & 0.344 & \textbf{0.334} & 0.572 &  &  &  & 0.97 \\
\multicolumn{12}{@{}l}{\quad\emph{\textsc{PEMS03}, FEDformer: noise level and placement}} \\
\textsc{PEMS03} & noise $\rho = 0.5$ & 107/107 & 1 & 0.663 & 0.332 & \textbf{0.286} & 0.767 & 18.9 & \textbf{27.4} & 19.2 & 0.98 \\
 & noise $\rho = 1$ & 107/107 & 1 & 0.650 & 0.319 & \textbf{0.290} & 0.743 & 19.1 & \textbf{27.3} & 19.1 & 0.99 \\
 & noise $\rho = 2$ & 107/107 & 1 & 0.618 & 0.312 & \textbf{0.291} & 0.735 & 19.6 & \textbf{27.3} & 18.9 & 0.99 \\
 & noise on target only & 107/107 & 1 & 0.651 & 0.327 & \textbf{0.280} & 0.763 & 19.2 & \textbf{27.6} & 19.1 & 0.99 \\
 & noise on trigger only & 107/107 & 1 & 0.656 & 0.319 & \textbf{0.286} & 0.758 & 19.0 & \textbf{27.3} & 19.1 & 0.99 \\
\midrule
\multicolumn{12}{@{}l}{\emph{Further knowledgeable-attacker combinations (levers as in Table~\ref{tab:knowledgeable}; corner: $5\%$ of channels, $0.5\%$ rate, sixteen-fold penalty, two-step trigger)}} \\
\textsc{PEMS03} & corner & 0/18 & 3 & 0.195 & 0.128 & 0.196 & \textbf{0.171} & 32.1 & 32.0 & \textbf{32.6} & 1.00 \\
\textsc{Weather} & O E T & 1/1 & 3 & 0.423 & 0.058 & \textbf{0.061} & 0.080 & 50.8 & \textbf{96.3} & 70.4 & 1.04 \\
\textsc{PEMS04} & C R & 11/31 & 0 & 0.297 & 0.172 & 0.253 & \textbf{0.245} & 36.0 & \textbf{39.1} & 35.3 & 1.00 \\
 & C R T & 31/31 & 0 & 0.280 & 0.170 & \textbf{0.186} & 0.223 & 33.1 & \textbf{41.5} & 35.8 & 1.00 \\
 & C R N E T & 31/31 & 0 & 0.272 & 0.168 & \textbf{0.183} & 0.216 & 33.5 & \textbf{41.8} & 36.2 & 1.00 \\
\bottomrule
\end{tabular}}
\end{table*}
\section{Clean-Data Cost and Sensitivity}
\label{app:sensitivity}

Table~\ref{tab:cleancost} lists the clean-data cost of Section~\ref{sec:e-utility} per model. Table~\ref{tab:sensitivity}
changes each setting of Table~\ref{tab:settings} in turn: recall stays between $0.996$ and $0.999$ ($0.998$ at the
defaults), the false flags vary more, most with fewer graph neighbours, and settings that bound minimum counts change no
flag. Varying the reconstruction settings keeps every series' reconstruction error within $9\%$ of its default value,
except a minimum of $25$ fit steps, which changes two series.

\paragraph{\textbf{Victim training loss}}\label{app:loss} Table~\ref{tab:losses} retrains the main FEDformer victims with squared error, the loss of Proposition~1, and with L1.
\textsc{SafeTime} leaves at most $0.018$ under \textsc{BackTime} and $0.044$ under \textsc{TDBA}, except \textsc{TDBA} on
\textsc{Weather} with squared error ($0.230$), where a clean-trained squared-error model also responds to the trigger
($0.648$).

\end{document}